\documentclass[10pt,twocolumn,letterpaper]{article}

\usepackage{cvpr}              

\usepackage{graphicx}
\usepackage{amsmath}
\usepackage{amssymb}
\usepackage{booktabs}
\usepackage{mmstyles}

\usepackage[pagebackref,breaklinks,colorlinks]{hyperref}

\usepackage[capitalize]{cleveref}
\crefname{section}{Sec.}{Secs.}
\Crefname{section}{Section}{Sections}
\Crefname{table}{Table}{Tables}
\crefname{table}{Tab.}{Tabs.}

\def\cvprPaperID{4080} 
\def\confName{CVPR}
\def\confYear{2022}

\begin{document}

\title{Coarse-to-fine Deep Video Coding with Hyperprior-guided Mode Prediction}

\author{
Zhihao Hu$^{1}$, \hspace{2mm}
Guo Lu$^{2}$, \hspace{2mm}
Jinyang Guo$^{3}$, \hspace{2mm}
Shan Liu$^4$, \hspace{2mm}
Wei Jiang$^4$, \hspace{2mm}
Dong Xu$^{3\dagger}$ \\
$^1$Beihang University, China \hspace{3mm}  $^2$Beijing Institute of Technology, China\\
$^3$The University of Sydney, Australia \hspace{3mm} $^4$Tencent, America\\ 
}


\maketitle

\begin{abstract}
{\let\thefootnote\relax\footnote{{$\dagger$~Dong Xu is the corresponding author.}}}
    The previous deep video compression approaches only use the single scale motion compensation strategy and rarely adopt the mode prediction technique from the traditional standards like H.264/H.265 for both motion and residual compression. In this work, we first propose a coarse-to-fine (C2F) deep video compression framework for better motion compensation, in which we perform motion estimation, compression and compensation twice in a coarse to fine manner. Our C2F framework can achieve better motion compensation results without significantly increasing bit costs. Observing hyperprior information (i.e., the mean and variance values) from the hyperprior networks contains discriminant statistical information of different patches, we also propose two efficient hyperprior-guided mode prediction methods. Specifically, using hyperprior information as the input, we propose two mode prediction networks to respectively predict the optimal block resolutions for better motion coding and decide whether to skip residual information from each block for better residual coding without introducing additional bit cost while bringing negligible extra computation cost. Comprehensive experimental results demonstrate our proposed C2F video compression framework equipped with the new hyperprior-guided mode prediction methods achieves the state-of-the-art performance on HEVC, UVG and MCL-JCV datasets.
\end{abstract}

\vspace{-8mm}
\section{Introduction}
\label{sec:intro}

Video compression systems are becoming more and more important for various practical applications due to the rapidly increasing demand for transmitting and storing huge amount of videos. While the conventional methods like H.264~\cite{wiegand2003overview}, H.265~\cite{sullivan2012overview} and the recent standard H.266~\cite{sullivan2020versatile} have achieved promising results based on different hand-crafted techniques, they cannot be end-to-end optimized by using large-scale video datasets.

Recently, a large number of deep video compression works~\cite{lu2019dvc,abdelaziz2019neural,Agustsson2020space,hu2021fvc} have been proposed (see Section~\ref{sec:relatedworks} for more details), and most of them follow the hybrid coding framework~\cite{wiegand2003overview,sullivan2012overview,sullivan2020versatile}, in which both motion compensation and residual compression modules are used to reduce the spatio-temporal redundancy. Therefore, two aspects are critical when designing new deep video codecs: 1) how to generate more accurate motion information for better motion compensation and 2) how to design more effective motion compression and residual compression approaches.

The state-of-the-art learning based video compression methods~\cite{lu2019dvc,Agustsson2020space,hu2021fvc} only use the single scale motion estimation and compensation strategy. Considering that the motion patterns in videos may be complex, these single-scale deep video codecs may not work well for compressing videos from complex scenarios with significant motion patterns. Motivated by the successful applications of the coarse-to-fine strategy for various tasks (\eg, optical flow estimation~\cite{ranjan2017optical} and video super-resolution~\cite{Wang2019EDVRVR}), in this work, we first propose a new coarse-to-fine deep video compression framework by adopting a two-stage motion compensation strategy to better generate the predicted feature. At the coarse level, given the low-resolution features from the reference frame and the current frame, we perform motion estimation to produce the low-resolution offset features, which are then compressed after using the motion compression module. After upsampling the reconstructed offset features, we further perform coarse-level motion compensation to wrap the high-resolution reference feature as the intermediate predicted feature. Based on this intermediate predicted feature and the high-resolution feature from the current frame, we perform these major operations (\ie, motion estimation, compression, and compensation) again at the fine level to additionally warp this intermediate predicted feature for better motion compensation. Our two-stage coarse-to-fine motion compensation strategy can generate better predicted feature for the subsequent residual compression module without significantly increasing the bit cost, which leads to better video compression performance.

To further improve video compression performance, we also propose two efficient mode prediction methods for both motion compression and residual compression, which are motivated by the success of the rate-distortion (RD) optimization based mode prediction methods in the traditional codecs~\cite{wiegand2003overview,sullivan2012overview,sullivan2020versatile} and the recent work~\cite{hu2020improving} (see Section~\ref{subsec:imrelate} for more details). 
Instead of using the computationally expensive RD optimization technology as in \cite{hu2020improving}, in this work, we propose to train two prediction networks for coding mode prediction, which bring negligible extra computational cost and can thus support more types of coding modes. Specifically, we use discriminant hyperprior information (\ie, the mean and variance values from the hyperprior network~\cite{minnen2018joint}) as the the input of the mode prediction networks as it represents the statistical characteristics of different patches and it does not introduce any additional bit cost. Our proposed mode prediction network can be readily used to adaptively select the optimal resolution of each block in motion compression or decide whether to skip residual information from each block in residual compression. 

Our contributions are summarized as follows: (1) We propose a simple and strong C2F deep video compression framework by performing two-stage motion compensation in a coarse-to-fine fashion. (2) We propose two hyperprior-guided mode prediction methods, in which we learn two mode prediction networks by using discriminant hyperprior information as the input. Our hyperprior-guided mode prediction methods do not introduce any additional bit cost, bring negligible computational cost, and can be readily used to predict the optimal coding modes (\ie, the optimal block resolution for motion coding and the ``skip"/``non-skip" mode for residual compression). (3) Comprehensive experiments on the HEVC, UVG and MCL-JCV datasets demonstrate our C2F framework equipped with the newly proposed hyperprior-guided mode prediction methods achieves comparable video compression performance with H265(HM)~\cite{HM} in terms of PSNR and generally outperforms the latest standard VTM~\cite{VTM} in terms of MS-SSIM.

\begin{figure*}[t]
\centering
\includegraphics[width=0.97\linewidth]{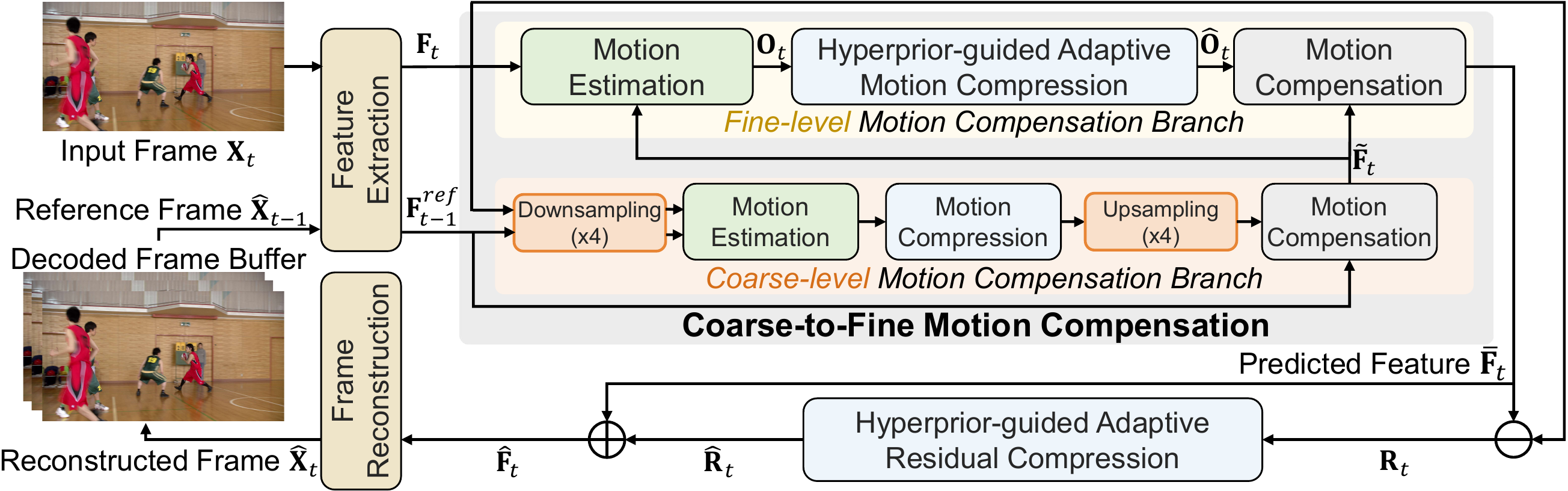}
\vspace{-3mm}
    \caption{Overview of our proposed video compression framework. Taking the input frame $\mX_t$ to be compressed and the reference frame $\hat{\mX}_{t-1}$ from the decoded frame buffer, we first perform the Feature Extraction operation to generate the input feature $\mF_t$ and the reference feature $\mF_{t-1}^{ref}$. Then the two-stage \textbf{Coarse-to-Fine Motion Compensation} module is used to compensate the reference feature in a coarse-to-fine fashion and generate the predicted feature $\Bar{\mF}_t$, in which a hyperprior-guided adaptive motion compression method is proposed for better motion compression. Last, the hyperprior-guided adaptive residual compression module is proposed to compress the residual feature $\mR_t$ between the input feature $\mF_t$ and the predicted feature $\Bar{\mF}_t$. After the frame reconstruction module, we produce the reconstructed frame $\hat{\mX}_{t}$, which is then stored in the decoded frame buffer. \vspace{-6mm}}
\label{fig:overview}
\end{figure*}

\vspace{-1mm}
\section{Related Work}
\label{sec:relatedworks}
\vspace{-1mm}

\subsection{Image and Video Compression}
\label{subsec:imrelate}
\vspace{-1mm}

To reduce the transmission and storage requirement from huge volumes of image/video data, different image and video compression standards such as JPEG~\cite{wallace1992jpeg}, JPEG2000~\cite{taubman2002jpeg2000}, H.264~\cite{wiegand2003overview}, H.265~\cite{sullivan2012overview} and H.266~\cite{sullivan2020versatile} were proposed. As these methods rely on hand-crafted techniques like DCT, they cannot be end-to-end optimized with other networks designed for various machine vision tasks (\eg, object detection).

Recently, a few learning-based image compression~\cite{toderici2015variable,toderici2017full,theis2017lossy,balle2016end,balle2018variational,minnen2018joint,cheng2019learning,cheng2020learned,minnen2020channel,chen2022exploiting} and video compression~\cite{wu2018video,lu2019dvc,rippel2019learned,abdelaziz2019neural,lombardo2019deep,lu2020anendtoend,lu2020content,Agustsson2020space,feng2020learned,mentzer2021towards,yang2020learning,liu2020mlvc,rippel2021elf,chen2022lsvc} methods have achieved promising compression performance. For example, Ball{\'e} \etal~\cite{balle2018variational} proposed to use the hyperprior network to reduce the entropy and save bits for better image compression. For video compression, the work FVC~\cite{hu2021fvc} performed all major operations in the feature space for better motion compensation and residual compression. However, the state-of-the-art video compression approaches~\cite{lu2019dvc,Agustsson2020space,hu2021fvc} only use the single scale motion compensation strategy. In contrast to these works~\cite{lu2019dvc,Agustsson2020space,hu2021fvc}, we propose a new deep video compression framework, in which we perform motion compensation twice in a coarse-to-fine fashion.

While the coding mode prediction technology is commonly employed in the traditional codes H.264/H.265, it is rarely used in deep video compression approaches except the recent work RaFC~\cite{hu2020improving}, in which the rate-distortion (RD) optimization technology is used for the resolution-adaptive motion coding. However, there are two drawbacks in \cite{hu2020improving}. First, it is computationally expensive to go through the whole network to calculate the RD value for each coding mode and thus it cannot support a large number of coding modes (\eg, only three coding modes are supported in ~\cite{hu2020improving}). Second, additional bits are required for encoding the predicted coding modes (\ie, the learnt binary masks representing the coding modes at different blocks).
In contrast to \cite{hu2020improving}, we propose to learn two networks for coding mode prediction by using discriminant hyperprior information as the input. Our approach does not introduce any additional bit costs, and brings negligible computational costs, which can support more types of coding modes and thus achieves better compression performance.

\subsection{Coarse-to-fine Strategy in Computer Vision}
The coarse-to-fine strategy is commonly used for various computer vision tasks including optical flow estimation~\cite{ranjan2017optical,hui2018liteflownet,sun2018pwc} and super-resolution~\cite{Wang2019EDVRVR}. Using the pyramid structure, the optical flow estimation method~\cite{ranjan2017optical} adopted the coarse-to-fine strategy to extract more accurate motion information. The video super-resolution method~\cite{Wang2019EDVRVR} also adopted the coarse-to-fine alignment strategy to generate better high-resolution frames.
However, how to use the coarse-to-fine strategy for video compression is not explored in the existing video compression methods.
In contrast to \cite{lu2019dvc,Agustsson2020space,hu2021fvc}, in this work, we propose a new coarse-to-fine video compression framework, in which our approach aims to address a different major challenge, namely, how to improve the motion compensation results without significantly increasing the bit cost.

\vspace{-1mm}
\section{Method}
\label{sec:method}

\vspace{-2mm}
\subsection{overview}
\label{subsec:overview}

Given the input video sequence $\mX=\{\mX_1, \mX_2, ..., \mX_{t-1}, \mX_t, ...\}$, video compression systems aim to reconstruct each video frame with high quality at any bit-rate. In this work, we directly use the existing image compression method~\cite{bellard2015bpg} to reconstruct the I frame and then employ the reconstructed previous frame as the reference frame for compressing the current frame. The overview of our proposed framework is shown in Fig.~\ref{fig:overview} and summarized as follows:

\textbf{Feature Extraction.}
Following the previous work FVC~\cite{hu2021fvc}, we first transform the input frame $\mX_t$ and the reference frame $\hat{\mX}_{t-1}$ into the input feature $\mF_t$ and the reference feature $\mF_{t-1}^{ref}$, respectively. The network structure of the feature extraction module is the same as that in FVC~\cite{hu2021fvc}, which consists of a convolution layer with stride 2 and a few residual blocks.

\textbf{Coarse-to-Fine Motion Compensation.}
In order to produce more accurate motion compensation results, we propose the two-stage coarse-to-fine motion compensation module. At the coarse level, we first generate two low resolution features by downsampling the input feature $\mF_t$ and the reference feature $\mF_{t-1}^{ref}$ and then we perform motion estimation, motion compression and the upsampling operation to generate the reconstructed coarse-level offset map, based on which we perform motion compensation to warp the reference feature $\mF_{t-1}^{ref}$ and eventually we generate the intermediate predicted feature $\tilde{\mF}_t$. Based on $\tilde{\mF}_t$ and the input feature $\mF_t$, in the fine-level motion compensation module, we perform three major operations including motion estimation, motion compression and motion compensation again at the fine level to generate the final predicted feature $\Bar{\mF}_t$. The network structure of the fine-level modules are the same as those in FVC~\cite{hu2021fvc}, except that we adopt the newly proposed hyperprior-guided adaptive motion compression module (see Section~\ref{subsec:HAMC} for more details), in which we learn a prediction network based on hyperprior information to decide the optimal block resolution for better motion coding.

\textbf{Hyperprior-guided Adaptive Residual Compression.}
The residual between the input feature $\mF_t$ and the final predicted feature $\Bar{\mF}_t$ is denoted by the residual feature $\mR_t$ and it will be compressed by the hyperprior-guided adaptive residual compression module (see Section~\ref{subsec:HAMC} for more details), in which based on hyperprior information we also learn a prediction network to predict the ``skip"/``non-skip" mode for better encoding residual features. Adding back the reconstructed residual feature $\hat{\mR}_t$ to the final predicted feature $\Bar{\mF}_t$, we produce the reconstructed feature $\hat{\mF}_t$.

\textbf{Frame Reconstruction.}
Feeding the reconstructed feature $\hat{\mF}_t$ into the frame reconstruction module that consists of a few residual blocks and a deconvolution layer~\cite{hu2021fvc}, we generate the reconstructed frame $\hat{\mX}_t$, which is then stored in the decoded frame buffer for processing the next frame.

\textbf{Entropy Coding.}
The encoded features from the coarse-level motion compression, the fine-level motion compression and the residual compression modules will be transformed into the bit-streams. During the training process, we use the bit-rate estimation network to predict the bit-rate. More details will be discussed in Section~\ref{subsec:others}.

\begin{figure}[t]
\centering
\includegraphics[width=0.85\linewidth]{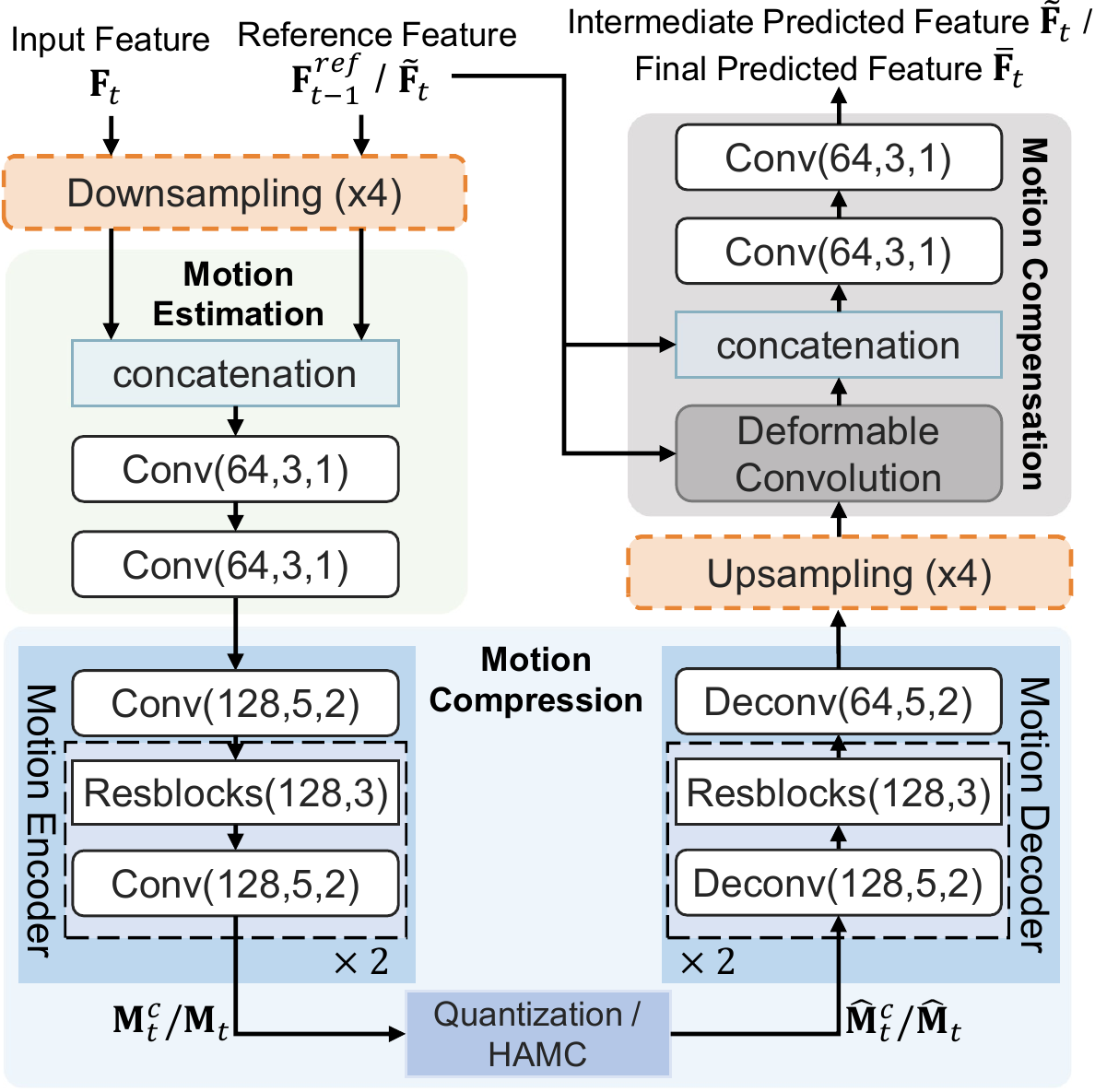}
\vspace{-3mm}
    \caption{Network structure of our proposed coarse-to-fine framework including three major modules (\ie, the motion estimation/compression/compensation modules). Coarse-level motion compensation takes the reference feature $\mF_{t-1}^{ref}$ and the input feature $\mF_t$ as the input to generate the intermediate predicted feature $\tilde{\mF}_t$, while fine-level motion compensation takes $\tilde{\mF}_t$ and $\mF_t$ as the input to generate the final predicted feature $\Bar{\mF}_t$. Note we use the dashed boxes for both ``upsampling" and ``downsampling" operations because the two modules are only used in the coarse-level motion compensation branch. ``HAMC" means Hyperprior-guided Adaptive Motion Compression. \vspace{-5mm}}
\label{fig:C2F}
\end{figure}

\vspace{-1mm}
\subsection{Coarse-to-Fine Motion Compensation}
\label{subsec:C2F-MC}
\vspace{-1mm}
To produce high quality features after motion compensation, we propose the C2F video compression framework by performing motion compensation in a coarse-to-fine fashion with only little computation and bit cost at the coarse level.  As shown in Fig.~\ref{fig:overview}, we first use the coarse-level motion compensation branch to generate the intermediate predicted feature $\tilde{\mF}_t$, and then take $\tilde{\mF}_t$ as the new reference feature to perform fine-level motion compensation. 

Specifically, as shown in Fig.~\ref{fig:C2F}, at the coarse-level motion compensation branch, the downsampling operation, which contains two convolution layers with stride 2, will transform the two features $\mF_{t-1}^{ref}$ and $\mF_t$ with the resolution of $H\times W$ into the coarse features with the resolution of $\frac{1}{4}H\times\frac{1}{4}W$. Based on the coarse features, we will go through the motion estimation module consisting of two convolutional layers, the motion compression module and the upsampling operation to generate the reconstructed coarse-level offset map with the resolution of $H\times W$. Here the coarse-level encoded motion feature $\mM_t^c$ will be quantized and used for entropy coding. Finally, we follow FVC~\cite{hu2021fvc} and use the deformable convolution~\cite{Dai2017DeformableCN} operation for feature space motion compensation, which takes the reconstructed offset map as the input to control the sampling location in the reference feature map. As in \cite{hu2021fvc}, we additionally concatenate the output of the deformable convolution layer with the reference feature and use two convolution layers to generate the intermediate predicted feature $\tilde{\mF}_t$.

The fine-level motion compensation branch is similar to the coarse-level motion compensation branch except that the input reference feature $\mF_{t-1}^{ref}$ is replaced by the intermediate predicted feature $\tilde{\mF}_t$ and the downsampling and upsampling modules are removed. In order to more effectively compress the motion information at the fine level, we propose a new hyperprior-guided adaptive motion compression method to adaptively compress the encoded motion feature $\mM_t$ into bit-stream, which will be discussed in Section~\ref{subsec:HAMC}.

By using the coarse-level motion compensation branch to first roughly compensate the reference feature $\mF_{t-1}^{ref}$ with little computation and bit costs, we can more accurately produce the predicted feature $\Bar{\mF}_t$ at the fine-level motion compensation branch, which eventually leads to better video compression performance.

\subsection{Hyperprior-guided Adaptive Motion Compression and Residual Compression}
\label{subsec:HAMC}

In order to more effectively compress motion information at the fine-level motion compensation branch, we propose the hyperprior-guided adaptive motion compression (HAMC) method to automatically predict the optimal block resolution for different spatial locations and different channels based on hyperprior information (\ie, the mean and variance values decoded from the hyperprior network), which can better compress motion information with no extra bits and negligible extra computation cost.


\begin{figure}[t]
\centering
\includegraphics[width=\linewidth]{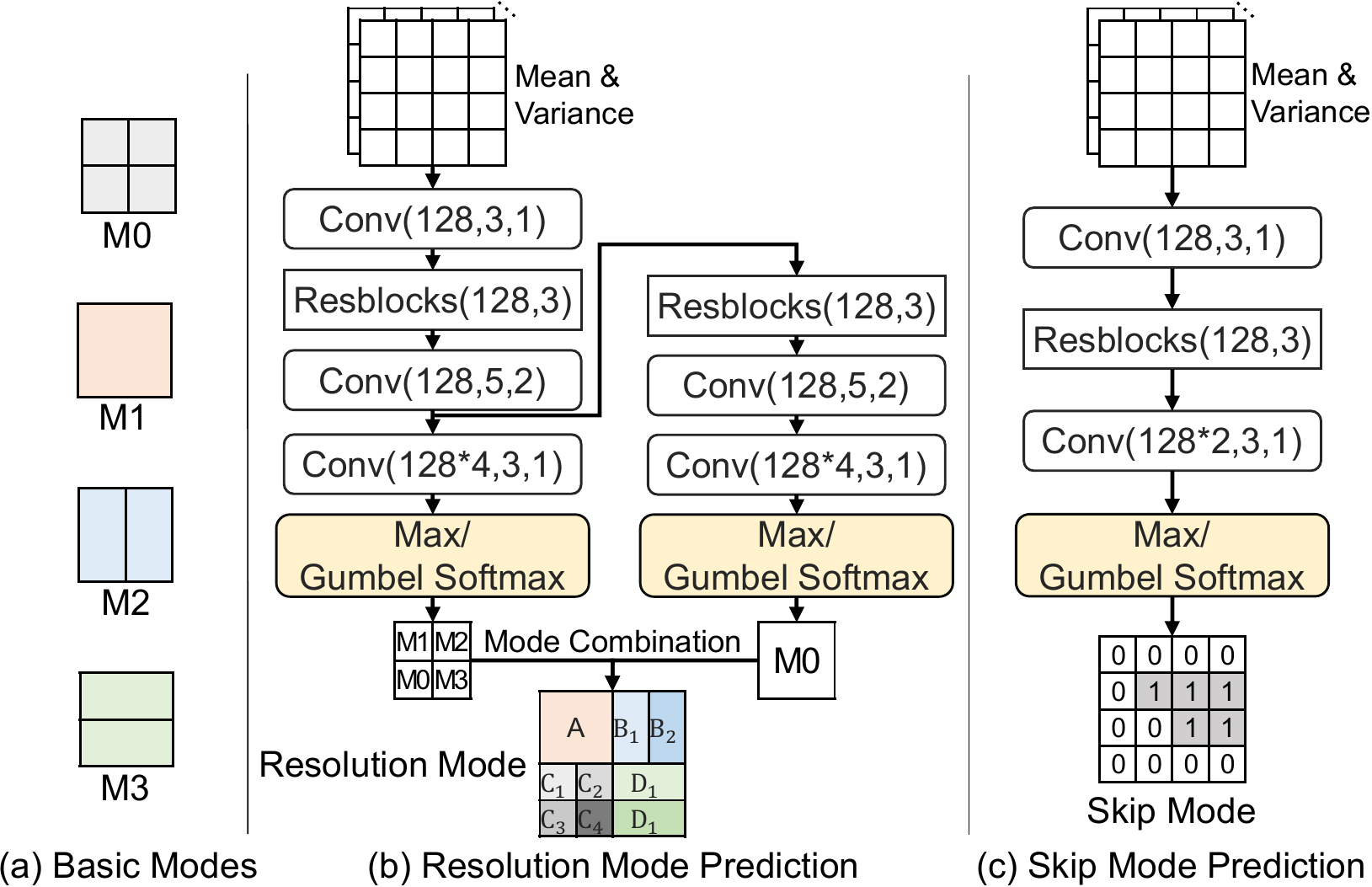}
\vspace{-6mm}
    \caption{
    (a) Four basic modes used in the resolution mode prediction network. (b) The resolution mode prediction network that predicts the optimal resolution for each block in the hyperprior-guided adaptive motion compression module and (c) the skip mode prediction network that predicts whether to skip residual information from each block in the hyperprior-guided adaptive residual compression module. For better illustration, we assume the size of the encoded motion/residual feature is $4\times4$ and only visualize the predicted mode from one channel. \vspace{-4mm}}
\label{fig:mp}
\end{figure}

Our method aims to decide the optimal partition mode for each target block, which can be predicted by using our resolution mode prediction network. For better illustration, below we assume the size of the encoded motion feature is $4\times4$. Specifically, our resolution mode prediction network consists of two branches, which predict the optimal mode for each $2\times2$ subblock and the overall $4\times4$ block, respectively. As shown in Fig.~\ref{fig:mp}(a), there are four basic modes for each $2\times2$ subblock (note similarly, we also have four basic modes for the target $4\times4$ block). Then the predicted modes from the two branches will be combined to generate a large number of possible coding modes for the target block. 
When the predicted mode for the target block is $\rm{M0}$, we additionally use the four modes predicted at each of the $2\times2$ subblocks to generate the updated coding mode for this target block. Otherwise, we directly use the mode predicted for this target block. 
An example is shown in Fig.~\ref{fig:mp}(b), as the mode predicted for the $4\times4$ target block is $\rm{M0}$, we use the four predicted modes [$\rm{M1}$, $\rm{M2}$; $\rm{M0}$, $\rm{M3}$] from the four corresponding $2\times2$ subblocks to generate the final predicted mode of this target block.

To predict the optimal mode, we propose a mode prediction network to automatically decide the resolution for each block based on hyperprior information, which represents the statistical information of each block.
As shown in Fig.~\ref{fig:mp}(b), we take the mean and variance values from the hyperprior decoder as the input of the mode prediction network to generate the confidence score of each mode. The channel number of the final convolution layer (\ie, before the gumbel softmax layer) is $128\times4$, which represents the confidence scores for 4 modes (shown in Fig.~\ref{fig:mp}(a)) over 128 channels for each $2\times2$ or $4\times4$ block. During the inference stage, we directly decide the optimal mode based on the maximum confidence score. However, the max operation is undifferentiable, so the whole network cannot be end-to-end optimized through back-propagation. To address this issue, during the training process, we adopt the Gumbel softmax strategy~\cite{Jang2017CategoricalRW} to decide the optimal mode as this Gumbel softmax module is differentiable, which thus enables end-to-end optimization for the whole network. Finally, after using the max/Gumbel softmax operation to generate the optimal mode for each $4\times4$ or $2\times2$ block and for each channel, we combine the predicted basic modes from each $4\times4$ or $2\times2$ block to generate the optimal resolution mode for this target block (see Fig.~\ref{fig:mp}(b)).

\begin{figure}[t]
\centering
\includegraphics[width=\linewidth]{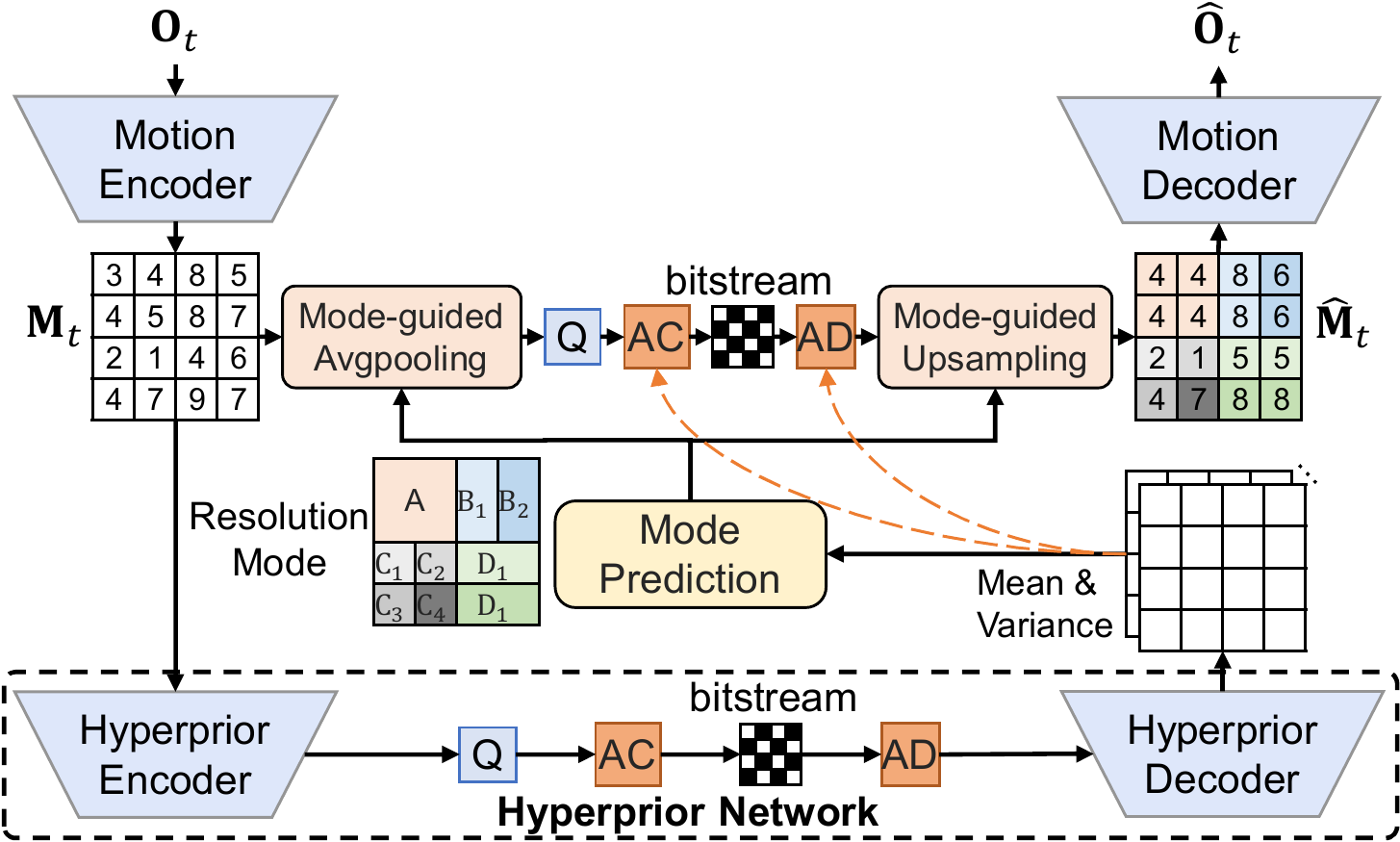}
\vspace{-6mm}
    \caption{The network structure of the hyperprior-guided adaptive motion compression method. AC and AD denote arithmetic coding and arithmetic decoding, respectively. Integers are used in $\mM_t$ and we only visualize one channel for better illustration. \vspace{-5mm}}
\label{fig:ende}
\end{figure}

The whole network structure of our proposed HAMC method is shown in Fig.~\ref{fig:ende}. We take the offset map $\mO_t$ (\ie, the output of the fine-level motion estimation module) as the input of the motion encoder to generate the encoded motion feature $\mM_t$. 
In our proposed HAMC, we predict the optimal resolution mode based on hyperprior information from the hyperprior network. According to the predicted resolution mode, we can more effectively transform the encoded motion feature into the bitstream. Taking subblock A (\ie, the top-left $2\times2$ subblock) in Fig.~\ref{fig:ende} as an example, we first perform the mode-guided avgpooling operation to average pool the four values $``3,4,4,5"$ in the left-top $2\times2$ subblock into only one value $``4"$, which is then quantized and transmitted as the bitstream by using the arithmetic coding (AC) operation. After the arithmetic decoding (AD) operation, we perform the mode-guided upsampling operation to generate the four values of $``4"$ in the left-top $2\times2$ subblock. Considering that we only transmit one value $``4"$ instead of four values $``3,4,4,5"$ to the decoder side (see the top-left $2\times2$ subblocks in $\mM_t$ and $\hat{\mM}_t$), our method uses much less bits. Similar operations are also performed for other three subblocks. 
Consequently, our proposed HAMC method can automatically select large block sizes in smooth areas with less significant motion patterns for bit-rate saving and use small block sizes for areas around moving object boundaries for achieving more accurate motion compensation results. 
In this way, we can effectively reduce the number of bits for transmitting the encoded motion feature $\mM_t$ without substantially degrading the quality of the reconstructed features, which leads to better compression results.


\textbf{Adaptive Residual Compression.}
We also propose the hyperprior-guided adaptive residual compression (HARC) method to more effectively compress the sparse residual information, in which we use the ``skip"/``non-skip" mode prediction network to predict whether to skip the residual information in each block.
As shown in Fig.~\ref{fig:mp}(c), the skip mode prediction network only use one-branch network instead of two branches as in HAMC. It also takes hyperprior information as the input to predict the ``skip"/``non-skip" mode for each entry at each channel of the encoded residual feature. As a result, the areas consisting of insignificant residual information will be predicted as the ``skip" mode for bit-rate saving, while the significant residual information will still be transmitted to the decoder side for better reconstruction result.



\begin{figure*}[t]
  \centering
  \begin{minipage}[c]{0.25\textwidth}
  \centering
    \includegraphics[width=\textwidth]{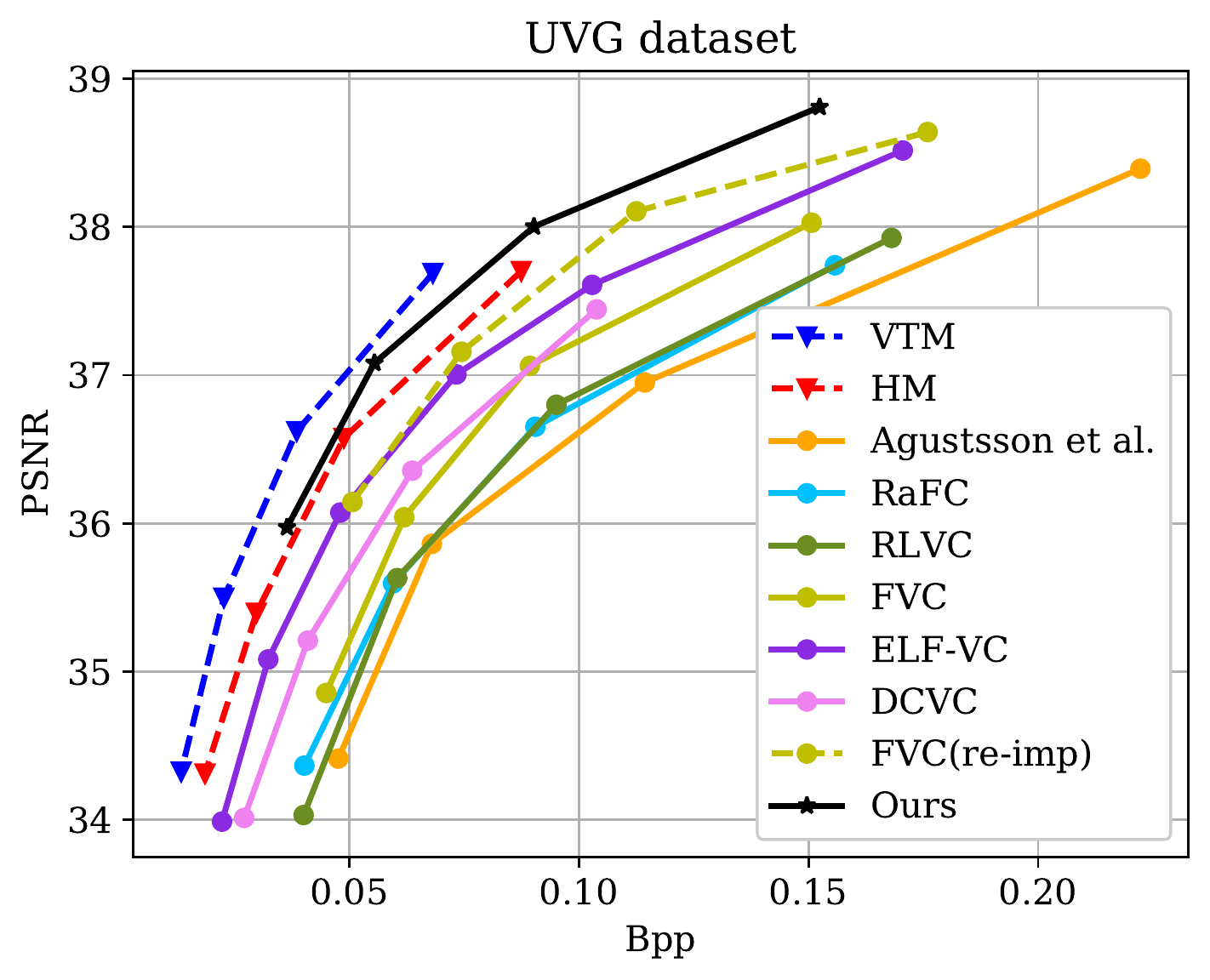}
  \end{minipage}%
  \begin{minipage}[c]{0.25\textwidth}
  \centering
    \includegraphics[width=\textwidth]{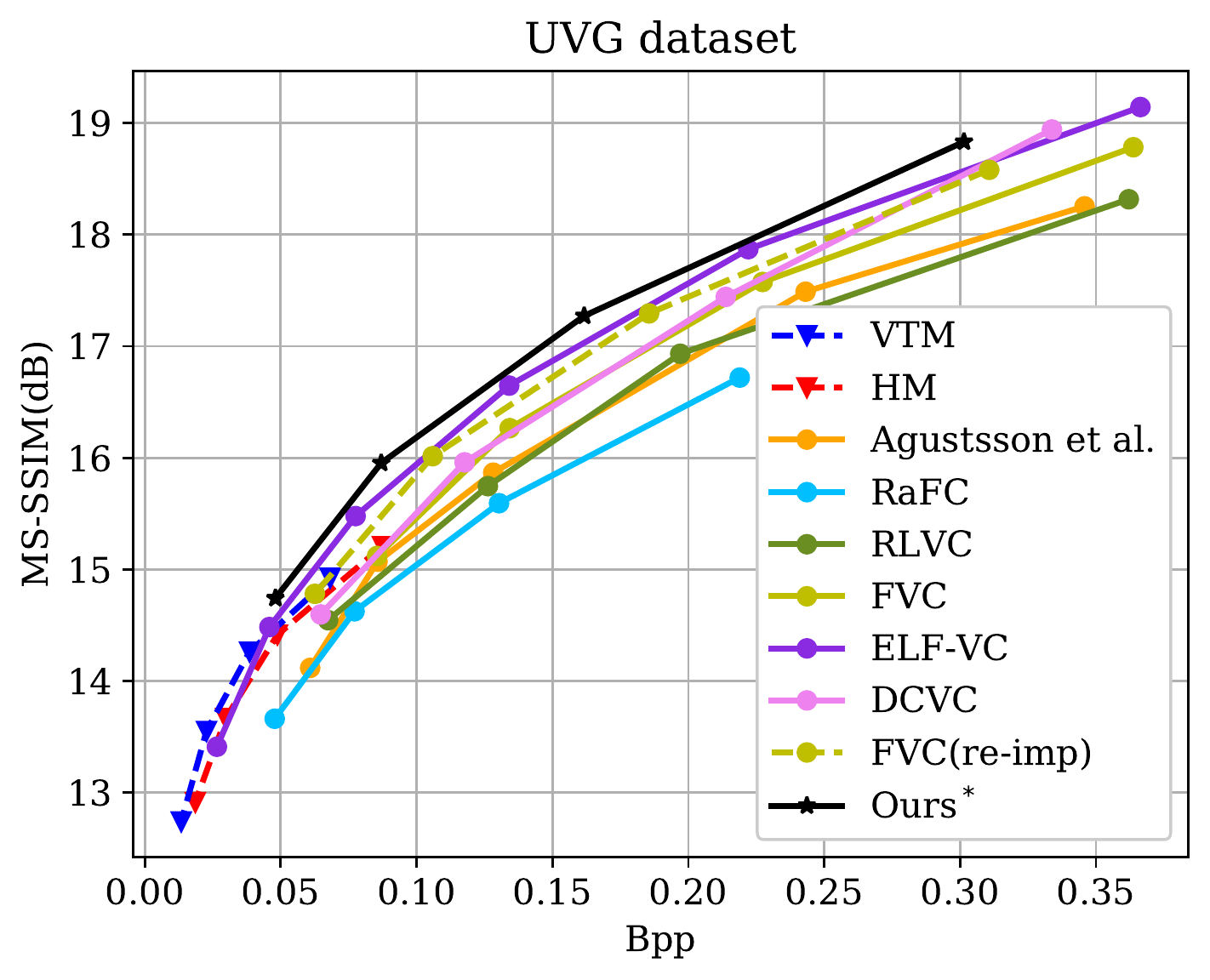}
  \end{minipage}%
  \begin{minipage}[c]{0.25\textwidth}
  \centering
    \includegraphics[width=\textwidth]{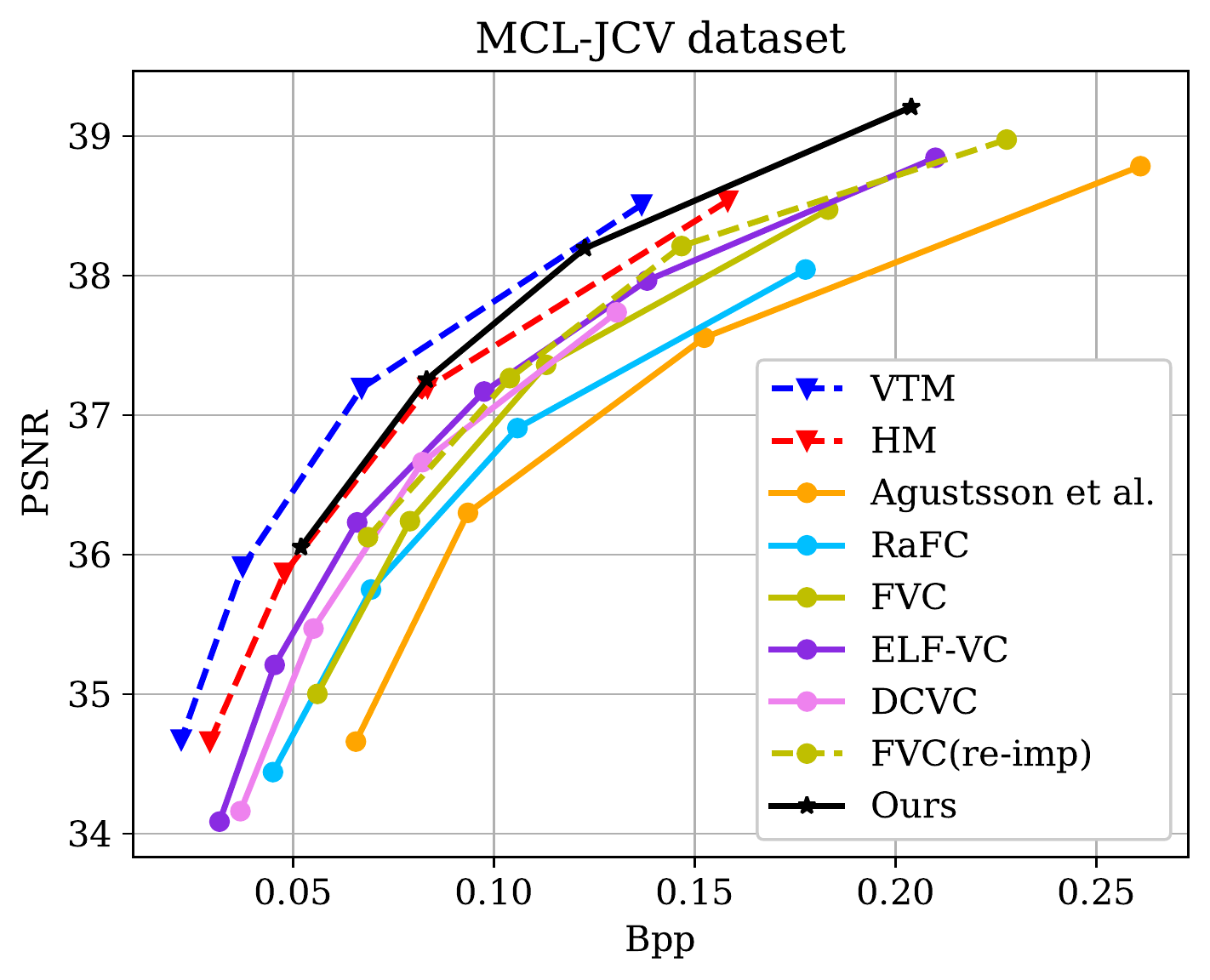}
  \end{minipage}%
  \begin{minipage}[c]{0.25\textwidth}
  \centering
    \includegraphics[width=\textwidth]{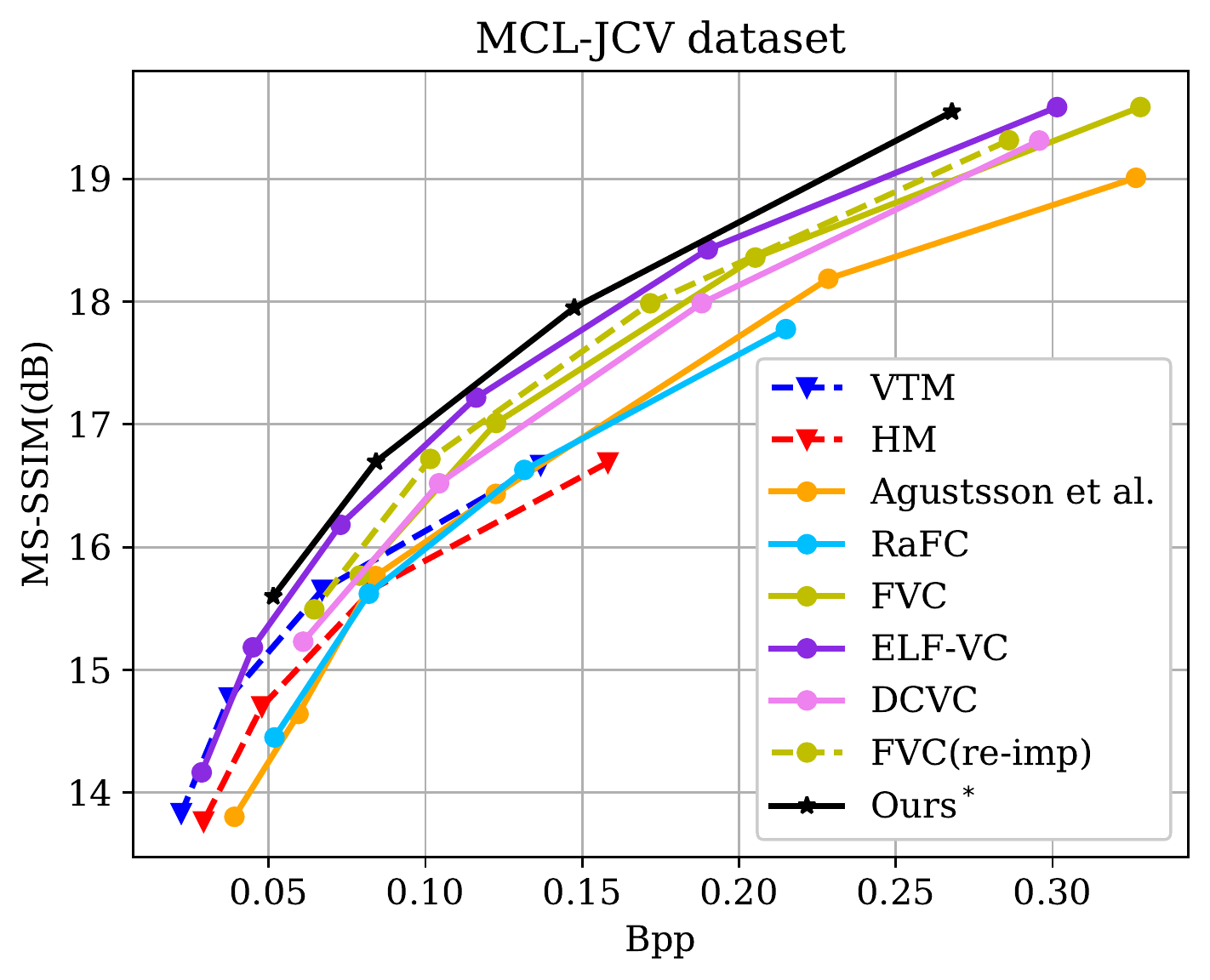}
  \end{minipage}
  \begin{minipage}[c]{0.25\textwidth}
  \centering
    \includegraphics[width=\textwidth]{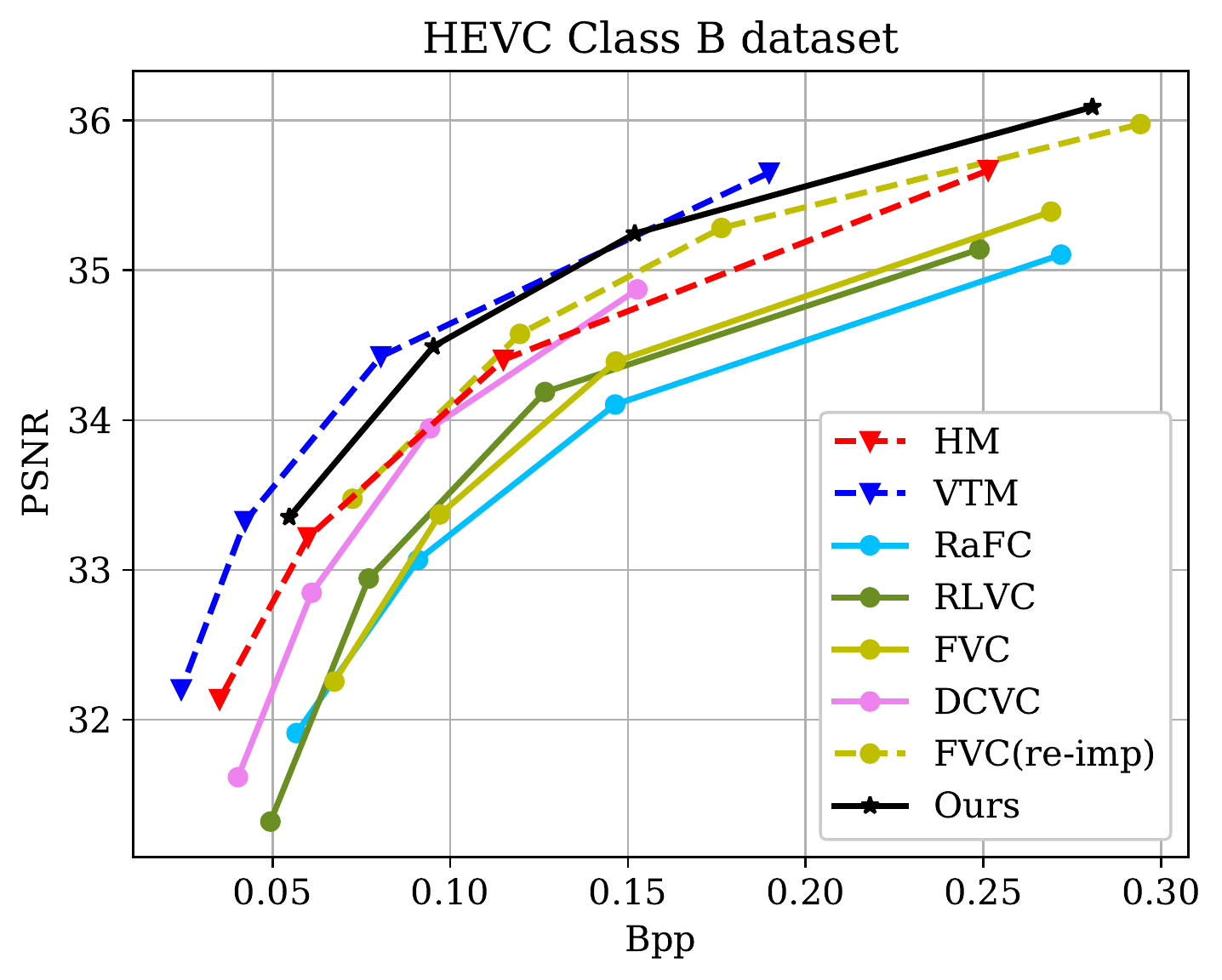}
  \end{minipage}%
  \begin{minipage}[c]{0.25\textwidth}
  \centering
    \includegraphics[width=\textwidth]{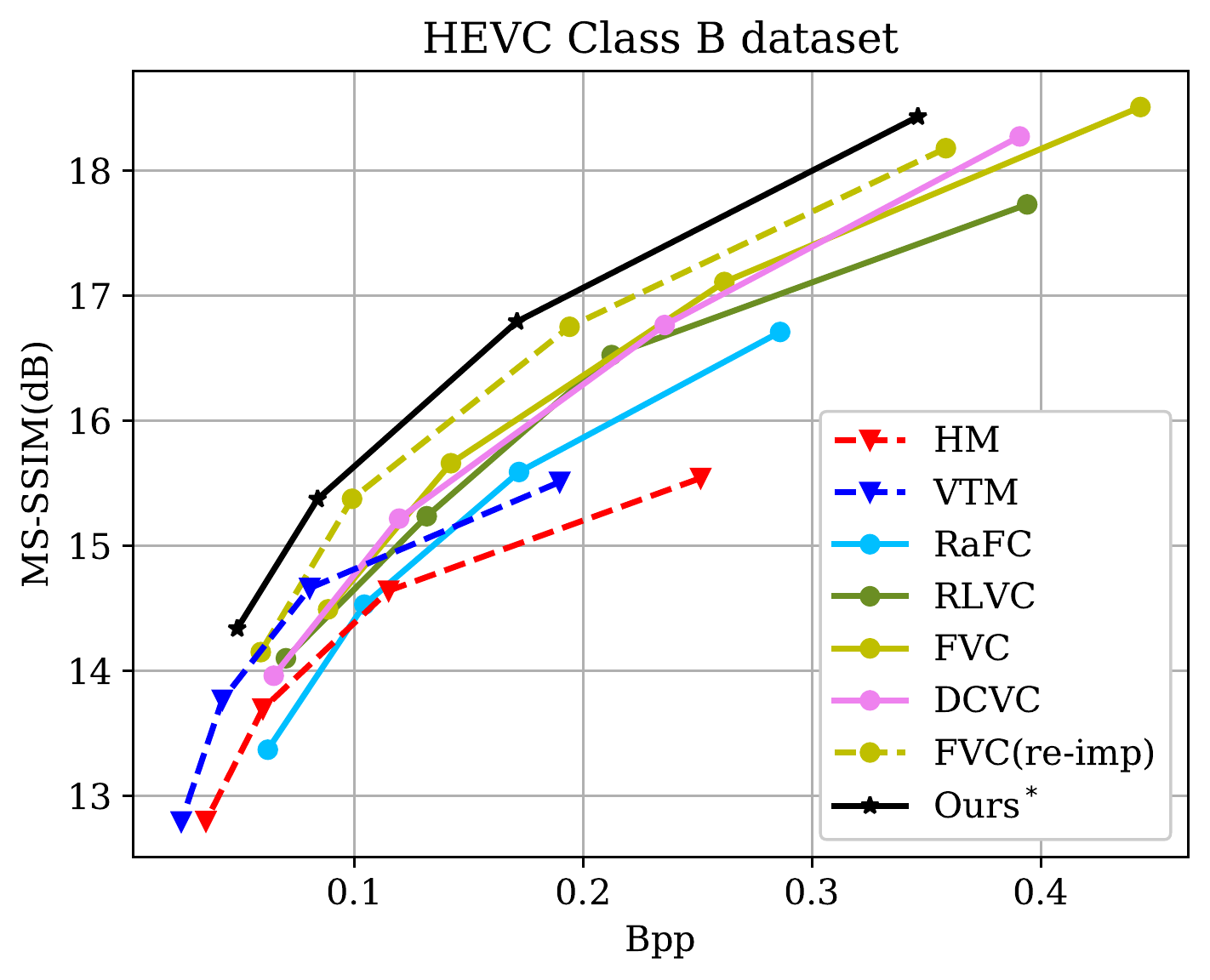}
  \end{minipage}%
  \begin{minipage}[c]{0.25\textwidth}
  \centering
    \includegraphics[width=\textwidth]{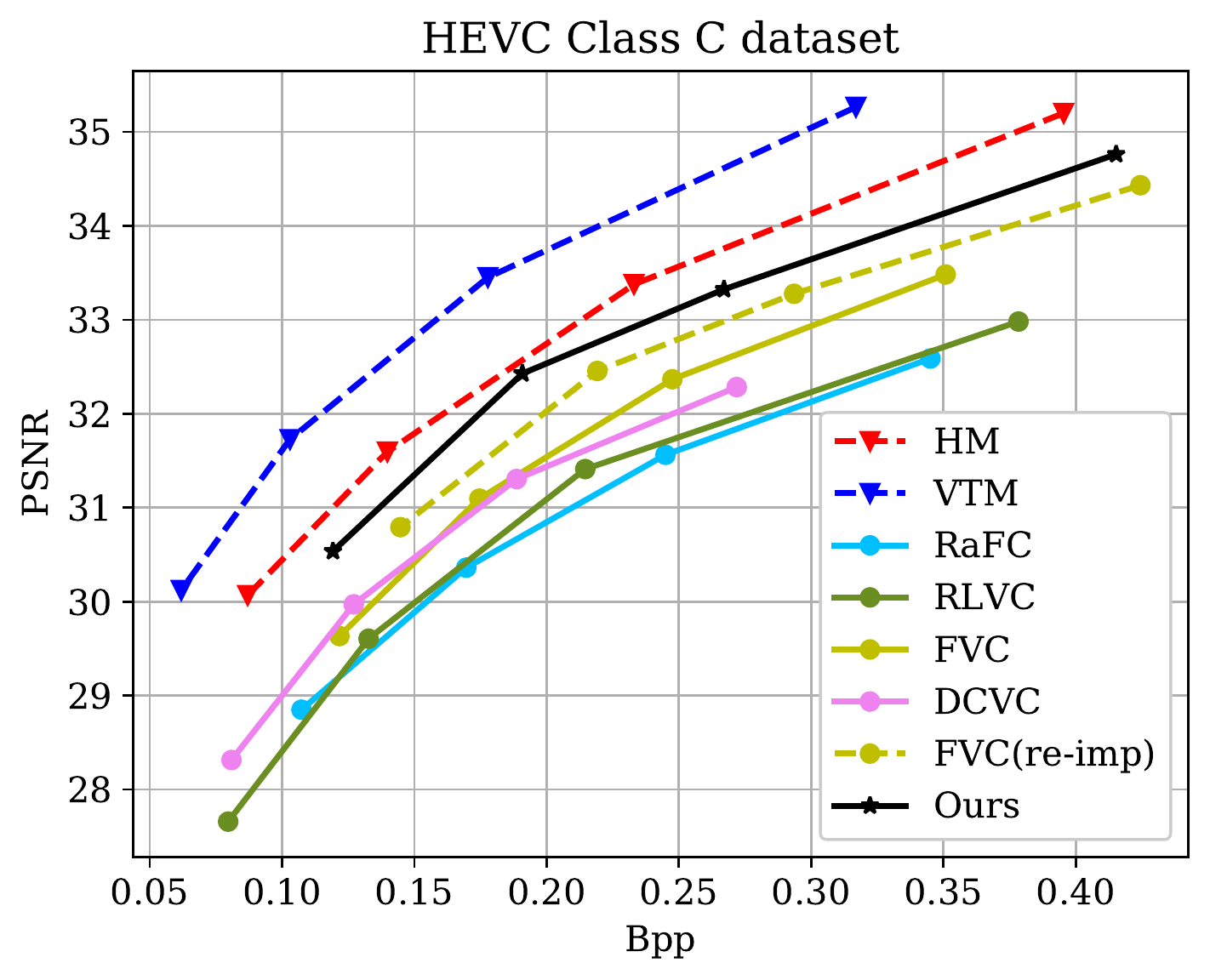}
  \end{minipage}%
  \begin{minipage}[c]{0.25\textwidth}
  \centering
    \includegraphics[width=\textwidth]{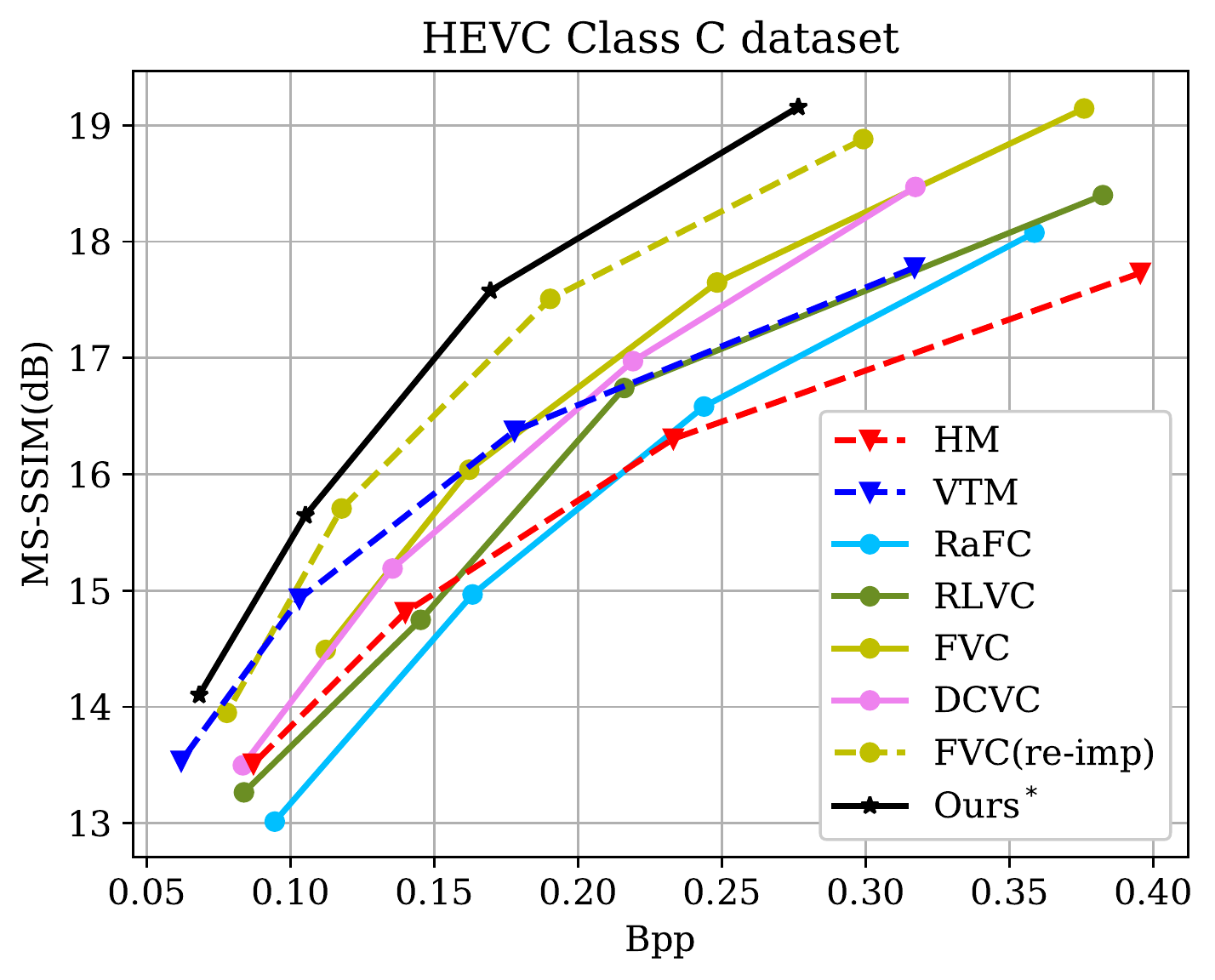}
  \end{minipage}
  \begin{minipage}[c]{0.25\textwidth}
  \centering
    \includegraphics[width=\textwidth]{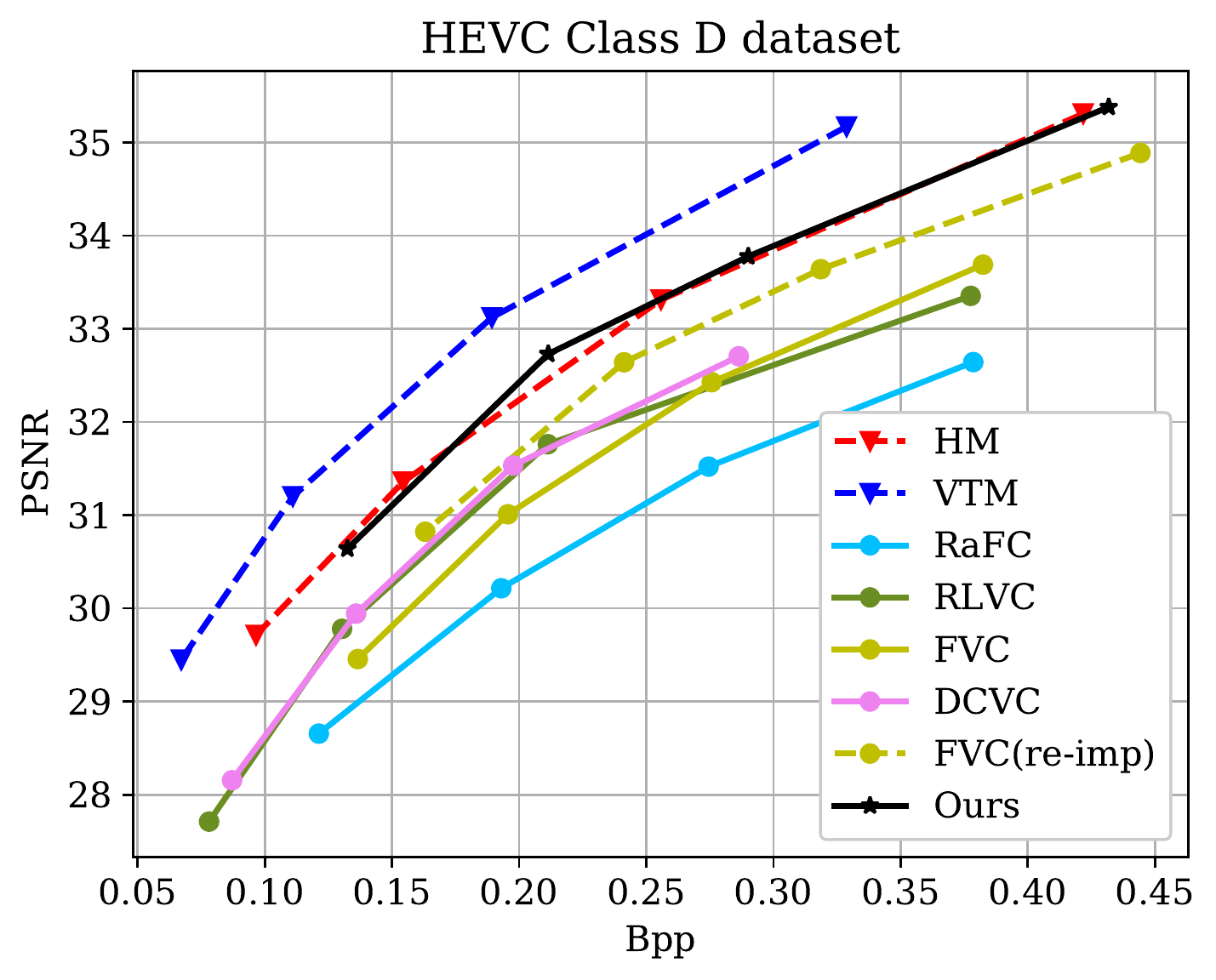}
  \end{minipage}%
  \begin{minipage}[c]{0.25\textwidth}
  \centering
    \includegraphics[width=\textwidth]{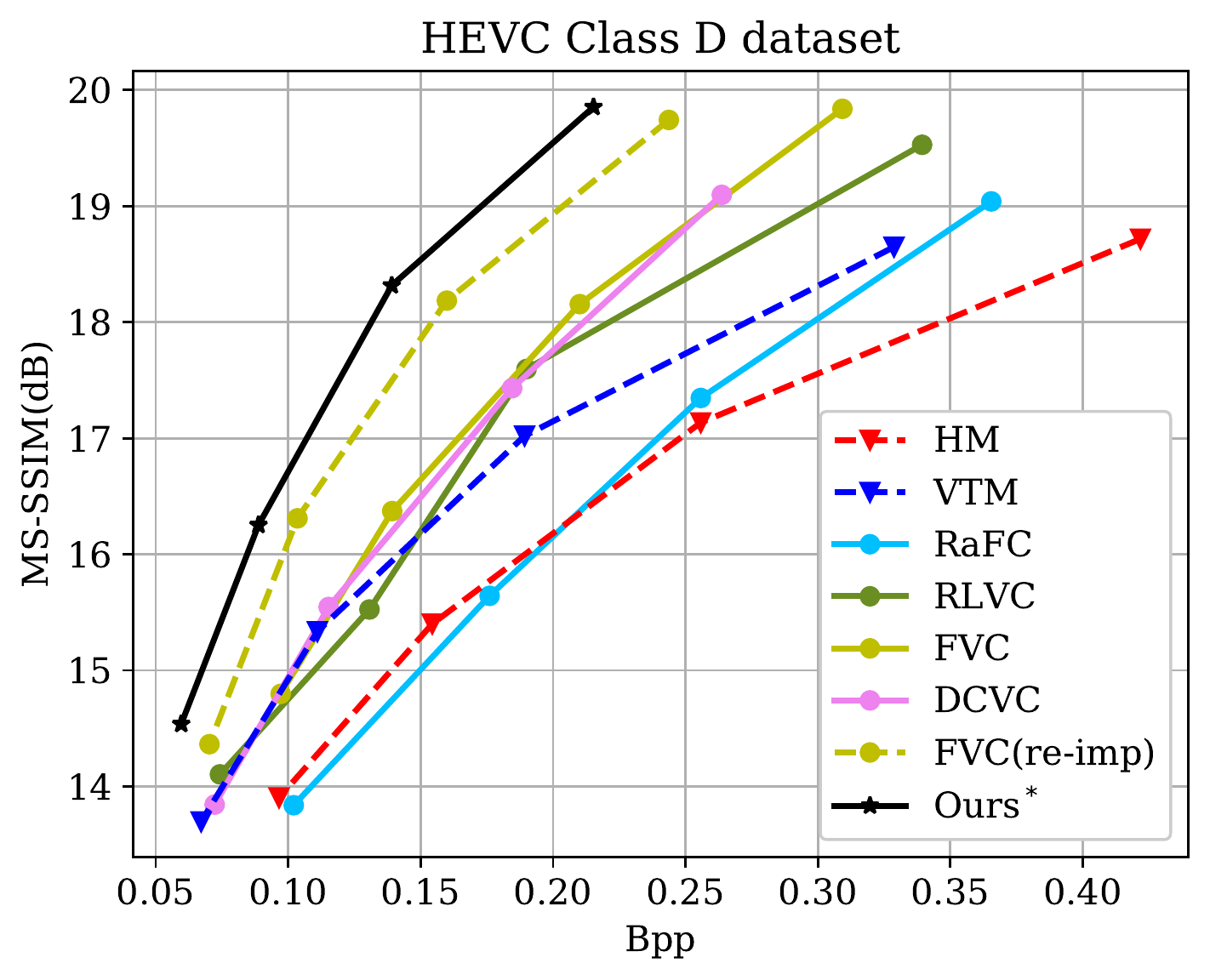}
  \end{minipage}%
  \begin{minipage}[c]{0.25\textwidth}
  \centering
    \includegraphics[width=\textwidth]{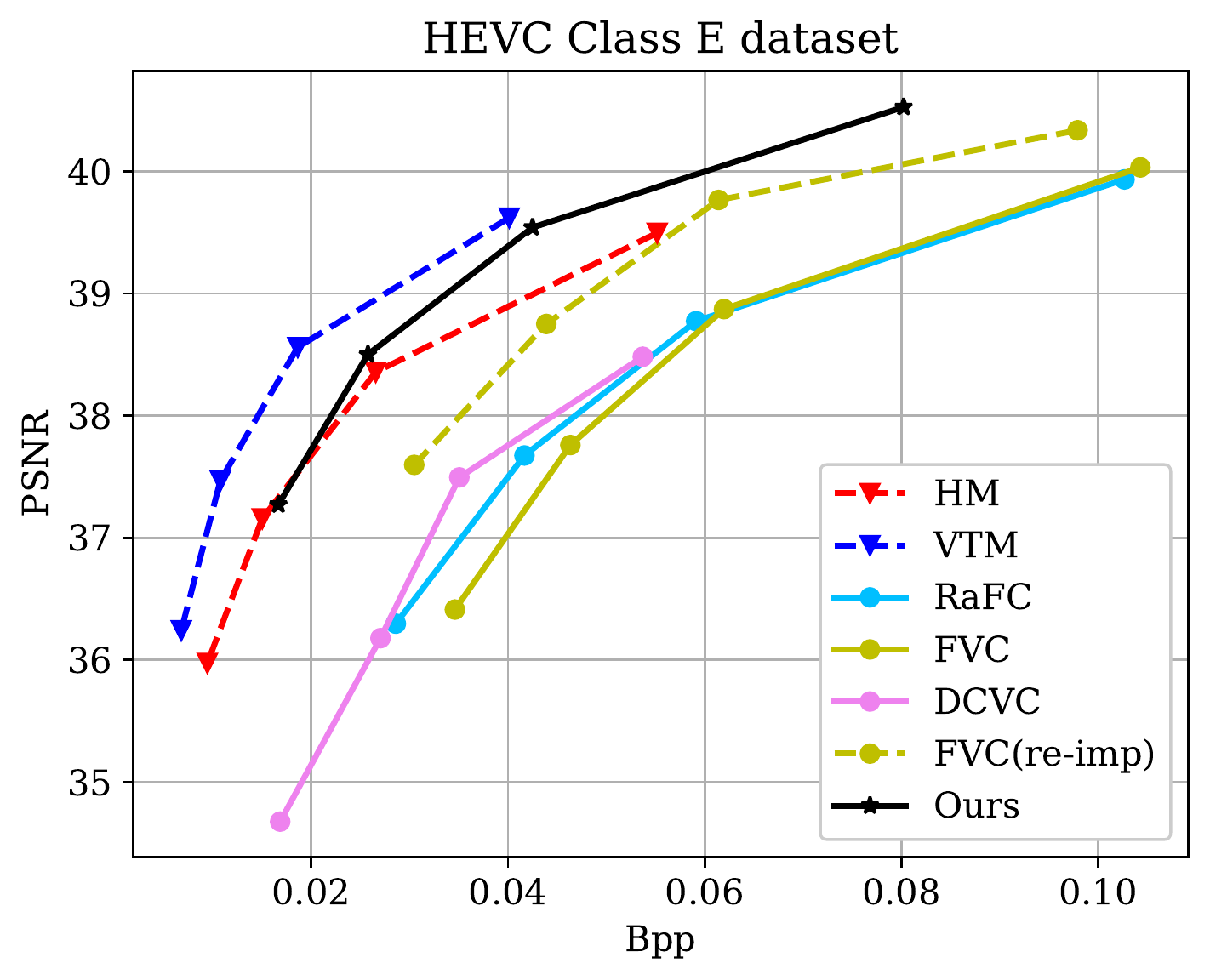}
  \end{minipage}%
  \begin{minipage}[c]{0.25\textwidth}
  \centering
    \includegraphics[width=\textwidth]{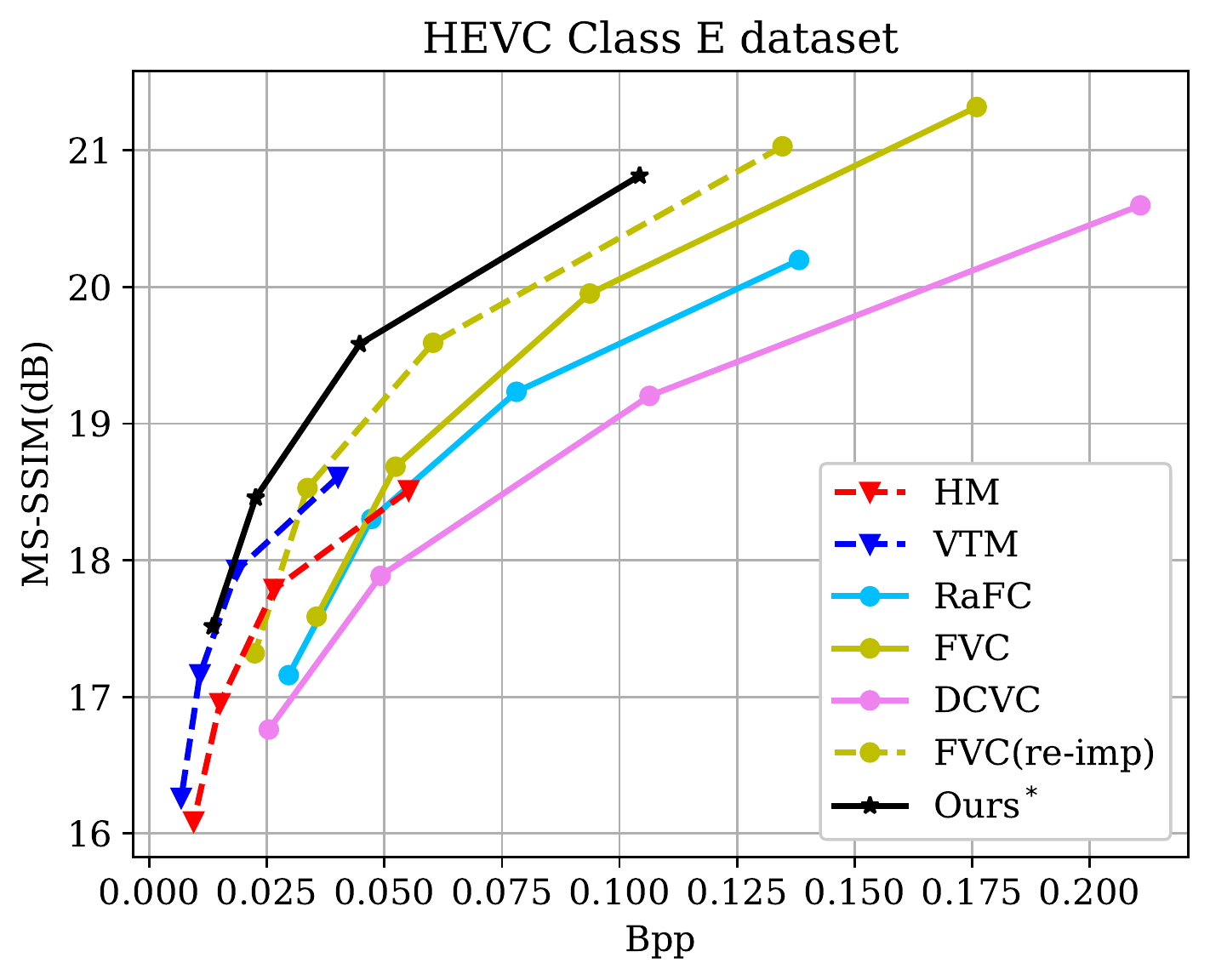}
  \end{minipage}
  \vspace{-4mm}
    \caption{The experimental results on the UVG, MCL-JCV, HEVC Class B, Class C, Class D and Class E datasets. \vspace{-4mm}}
  \label{fig:result}
\end{figure*}

\vspace{-1mm}
\subsection{Loss Function and Entropy Coding}
\label{subsec:others}
\vspace{-1mm}

The whole network is end-to-end optimized by minimizing the following rate-distortion loss,
\begin{equation}
\begin{aligned}
\label{eq:rd}
\mathcal{L} = \mathbb{H}(\hat{\mM}_t^c) + \mathbb{H}(\hat{\mM}_t) + \mathbb{H}(\hat{\mY}_t) + \lambda d(\mX_t, \hat{\mX}_t),
\end{aligned}
\end{equation}
where $\mathbb{H}(\cdot)$ denotes the number of bits for encoding the features including the quantized encoded coarse-level motion feature $\hat{\mM}_t^c$, the quantized encoded fine-level motion feature $\hat{\mM}_t$ and the quantized encoded residual feature $\hat{\mY}_t$. $d(\mX_t, \hat{\mX}_t)$ denotes the distortion between the input frame $\mX_t$ and the reconstructed frame $\hat{\mX}_t$. $\lambda$ is the hyper-parameter that controls the trade-off between the bit-rate and distortion.
During the training process, we adopt the bit-rate estimation network from \cite{minnen2018joint} without using the time-consuming auto-regressive model to estimate the bits for compressing $\hat{\mM}_t$ and $\hat{\mY}_t$. Considering that the resolution of coarse-level motion feature is relative small, we directly use the simple bit-rate estimation network in \cite{balle2016end} for estimating the bits for $\hat{\mM}_t^c$.

\vspace{-2mm}
\section{Experiments}
\label{sec:experiments}
\vspace{-1mm}
\subsection{Experimental Setup}
\label{subsec:expsetup}
\vspace{-1mm}

\textbf{Training Dataset.} 
Following the previous works~\cite{lu2019dvc,hu2021fvc}, we use the Vimeo-90K dataset~\cite{xue2019video} during the training stage.
This dataset contains 89,800 video sequences with each video sequence consisting of 7 consecutive frames with the resolution of $488\times256$. 
For data augmentation, the video sequences are random flipped and random cropped into $256\times256$ patches before feeding into the network.

\textbf{Testing Datasets.}
We evaluate our performance on multiple datasets including the HEVC~\cite{sullivan2012overview} Class B, C, D, E, UVG~\cite{UVGdataset} and MCL-JCV~\cite{wang2016mcl} datasets. The HEVC standard datasets~\cite{sullivan2012overview} contain different types of video sequences with various resolutions including $1920\times1080$ (Class B), $832\times480$ (Class C), $416\times240$ (Class D) and $1280\times720$ (Class E). The UVG dataset~\cite{UVGdataset} contains seven 1080p video sequences with high frame rate and the MCL-JCV dataset~\cite{wang2016mcl} contains thirty 1080p video sequences, which are widely used for learning-based video codec evaluation.

\textbf{Evaluation Metric.}
PSNR and MS-SSIM~\cite{wang2003multiscale} are used to evaluate the video quality. PSNR is the most popular metric for evaluating the video sequence distortion and MS-SSIM is commonly adopted for subjective visual quality evaluation. Bit per pixel (bpp) is used to evaluate the number of bits for compressing motion information and residual information.

\textbf{Implementation Details.}
We train our model in three stages. At the first stage, we use two consecutive frames including one I frame and one P frame to train our model for 2,000,000 steps without adopting both HAMC and HARC schemes. Then we extend the length of the training video sequence to 7 frames at the second stage for another 300,000 steps. Finally, we add our newly proposed HAMC and HARC schemes and train our complete model for 200,000 steps. The initial learning rate is set as 5e-5, which is decreased by 80\% at the 1,900,000th step and the 2,400,000th step. We set the batch size as 4 for the first stage and 2 for other stages. We use the Adam optimizer~\cite{kingma2014adam} based on PyTorch with CUDA support. 
We use the mean square error as the distortion loss for the PSNR results and additionally fine-tune the PSNR models by using MS-SSIM as the distortion loss for 100,000 steps to produce the MS-SSIM results.
When training our model on the machine with a single 2080TI GPU, it takes about 4.5 days, 2 days and 1.3 days for the first stage, the second stage and the final stage, respectively, and it costs 15 hours for fine-tuning the MS-SSIM results. 
In order to minimize the influence of the video length mismatch between the training sequence and the testing sequence, we additionally adopt the random shift and different distortion weight strategies as suggested in ~\cite{mentzer2021towards}.

\vspace{-1mm}
\subsection{Experimental Results}
\label{subsec:expresult}
\vspace{-1mm}

To evaluate the effectiveness of our proposed method, we compare our proposed method with the state-of-the-art learning-based methods including Agustsson~\etal~\cite{Agustsson2020space}, RaFC~\cite{hu2020improving}, RLVC~\cite{yang2021learning}, FVC~\cite{hu2021fvc}, DCVC~\cite{li2021deep} and ELF-VC~\cite{rippel2021elf}. Based on the same setting as our proposed method, we further provide the re-implementation results of FVC~\cite{hu2021fvc} without adopting the multi-frame feature fusion module, which is denoted by ``FVC(re-imp)" and used as our baseline method. For the conventional methods, we directly use the standard H.265(HM)~\cite{HM} and VTM~\cite{VTM} with the low delay P configuration for comparison. Different from the previous methods that use the commercial software FFmpeg to generate the results of x265, HM and VTM are the standard versions that can achieve much better performance but are extremely slow.

In order to fairly compare with the standard HM and VTM, we set the GoP size as 100 for all datasets and use BPG~\cite{bellard2015bpg} for I frame compression. 
To minimize the cumulative error, we also follow HM and VTM to use a better P-frame compression model for compressing the fourth frame of every 4 frames.
When using MS-SSIM for performance evaluation, our model is further fine-tuned by using the MS-SSIM loss as the distortion loss, which is denoted by $\rm{``Ours^*"}$.

As shown in Fig.~\ref{fig:result}, our method outperforms all other learning-based methods by a large margin in terms of PSNR. When compared with the recently proposed ELF-VC on the UVG dataset, our proposed method achieves 0.5dB improvement at 0.1bpp. Compared with the conventional method H.265(HM) in terms of PSNR, our method achieves better results on most datasets. VTM~\cite{VTM} is the latest video compression standard, which executes various hand-designed modules to achieve the current best performance and thus runs extremely slow (less than 0.001fps). Although the performance of VTM is better than our method in terms of PSNR, we observe that our results are close to VTM at high bit-rate on all high-resolution datasets (\ie, UVG, MCL-JCV, HEVC Class B and Class E). Besides, our method runs at 3.41fps, which is 3000x faster than VTM. Our method generally outperforms all baseline methods in terms of MS-SSIM.
Additionally, when using H.265(HM) as the anchor method, our average bit-rate saving over the HEVC Class B,C,D,E datasets is 4.58\%. 

\begin{figure}[t]
\centering
  \includegraphics[width=0.6\linewidth]{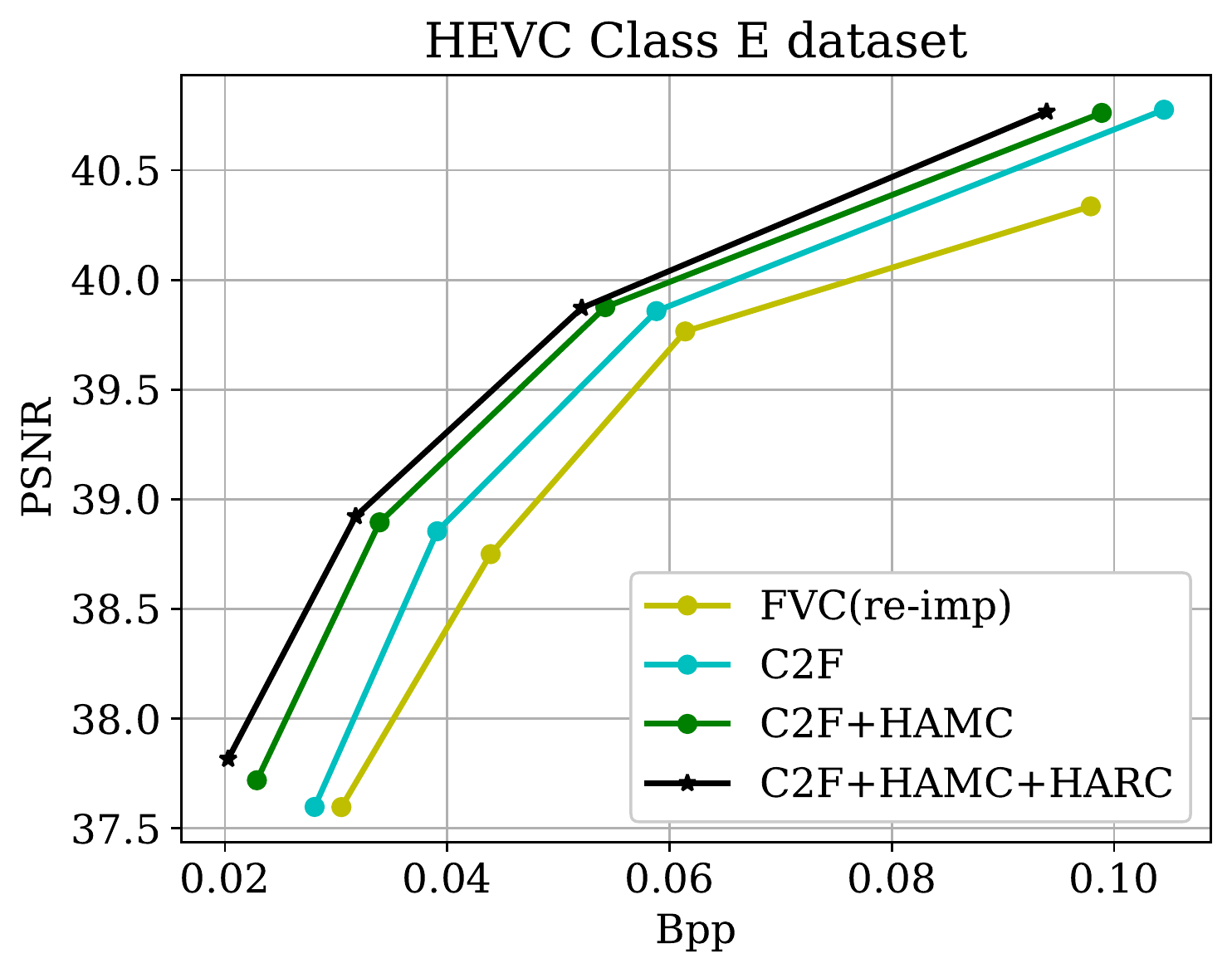}
  \vspace{-4mm}
  \caption{Ablation study on the HEVC Class E dataset. (1) \textbf{FVC(re-imp)}: Our re-implemented baseline method FVC~\cite{hu2021fvc}. (2) \textbf{C2F}: Our proposed coarse-to-fine video compression framework. (3) \textbf{C2F+HAMC}: Our proposed C2F framework equipped with the hyperprior-guided adaptive motion compression (HAMC) method. (4) \textbf{C2F+HAMC+HARC}: Our proposed C2F framework equipped with both HAMC and the hyperprior-guided adaptive residual compression (HARC) method. \vspace{-4mm}}
\label{fig:ablation}
\end{figure}

\vspace{-1mm}
\subsection{Ablation Study}
\label{subsec:ablation}
\vspace{-1mm}
The ablation study of our proposed method is shown in Fig.~\ref{fig:ablation}. We take FVC(re-imp) as our baseline method. When compared with FVC(re-imp), our coarse-to-fine framework C2F achieves 0.3dB improvement at 0.08bpp , which indicates the effectiveness of our proposed coarse-to-fine strategy for motion compensation. When comparing C2F+HAMC with C2F, the newly proposed hyperprior-guided adaptive motion compression (HAMC) method brings 0.6dB improvement at 0.03bpp. In addition, we observe that the coarse-to-fine strategy improves more at higher bit-rate, while HAMC achieves more improvement at lower bit-rate. One possible explanation is that the coarse-to-fine strategy aims to improve the quality of the predicted feature and it is thus more beneficial to use the C2F framework for generating high quality reference frame, while our HAMC scheme focuses on bit-rate saving and it can thus reduce more redundancy at lower bit-rate. Finally, our proposed coarse-to-fine framework equipped with both HAMC and the hyperprior-guided adaptive residual compression (HARC) methods achieves the best results and outperforms the baseline FVC(re-imp) by 1.2dB at 0.03bpp, which demonstrates the effectiveness of our proposed C2F framework and two new methods HAMC and HARC.


\vspace{-1mm}
\subsection{Model Analysis}
\label{subsec:analysis}
\vspace{-1mm}


\begin{figure}[t]
\centering
  \includegraphics[width=0.95\linewidth]{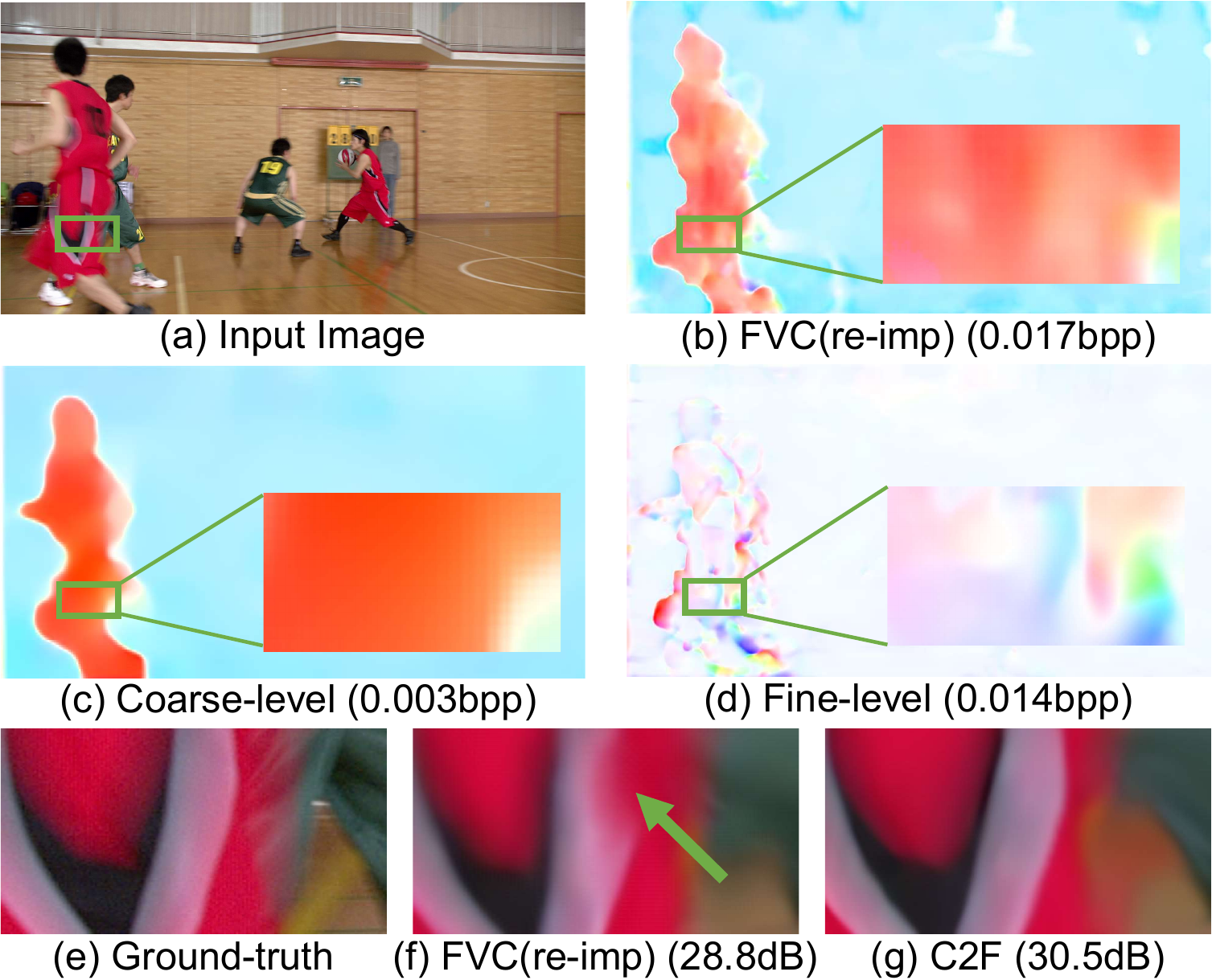}
  \vspace{-2mm}
  \caption{Visualization of (a) the input frame, (b) the offset map from the baseline method FVC(re-imp) using single scale motion compensation and (c) the coarse-level offset map and (d) the fine-level offset map from our coarse-to-fine framework on the HEVC Class B dataset. We also provide (e) one ground-truth patch and its corresponding motion compensation results by using (f) the baseline method FVC(re-imp) and (g) our C2F framework. Bpps for compressing the corresponding offset maps and PSNRs of the corresponding motion compensation results are also reported.\vspace{-6mm}}
\label{fig:visC2F}
\end{figure}

\textbf{Visualization of Coarse-to-fine Motion Compensation.}
To verify the effectiveness of our proposed coarse-to-fine strategy for motion compensation, we take the 5th reconstructed P frame from the first video of the HEVC Class B dataset to visualize both motion information and the corresponding motion compensation results. For fair comparison, here we use the same fine-level motion compression network as FVC(re-imp) without adopting the HAMC method in our C2F framework. It is observed that our coarse-level offset map (see Fig.~\ref{fig:visC2F}(c)) consists of coarse patch-level motion information, which is easy to be compressed and only costs 0.003bpp. We also observe that the fine-level offset map (see Fig.~\ref{fig:visC2F}(d)) can capture more detailed pixel-level motion information than the coarse-level offset map, which can better handle more complex motion patterns. As a result, our proposed two-stage coarse-to-fine motion compensation framework can achieve more accurate motion compensation result (see Fig.~\ref{fig:visC2F}(g)). When compared with FVC(re-imp) (see Fig.~\ref{fig:visC2F}(f)), the PSNR of our method is improved by 1.7dB with similar bit-rate cost, which demonstrates that our coarse-to-fine strategy can significantly improve the motion compensation performance. Additionally, when compared with the baseline method FVC(re-imp) on the HEVC Class E dataset, on average the motion compensation result after using our C2F framework is improved by 1.14dB, while costing similar bpps.

\begin{figure}[t]
\centering
  \includegraphics[width=0.95\linewidth]{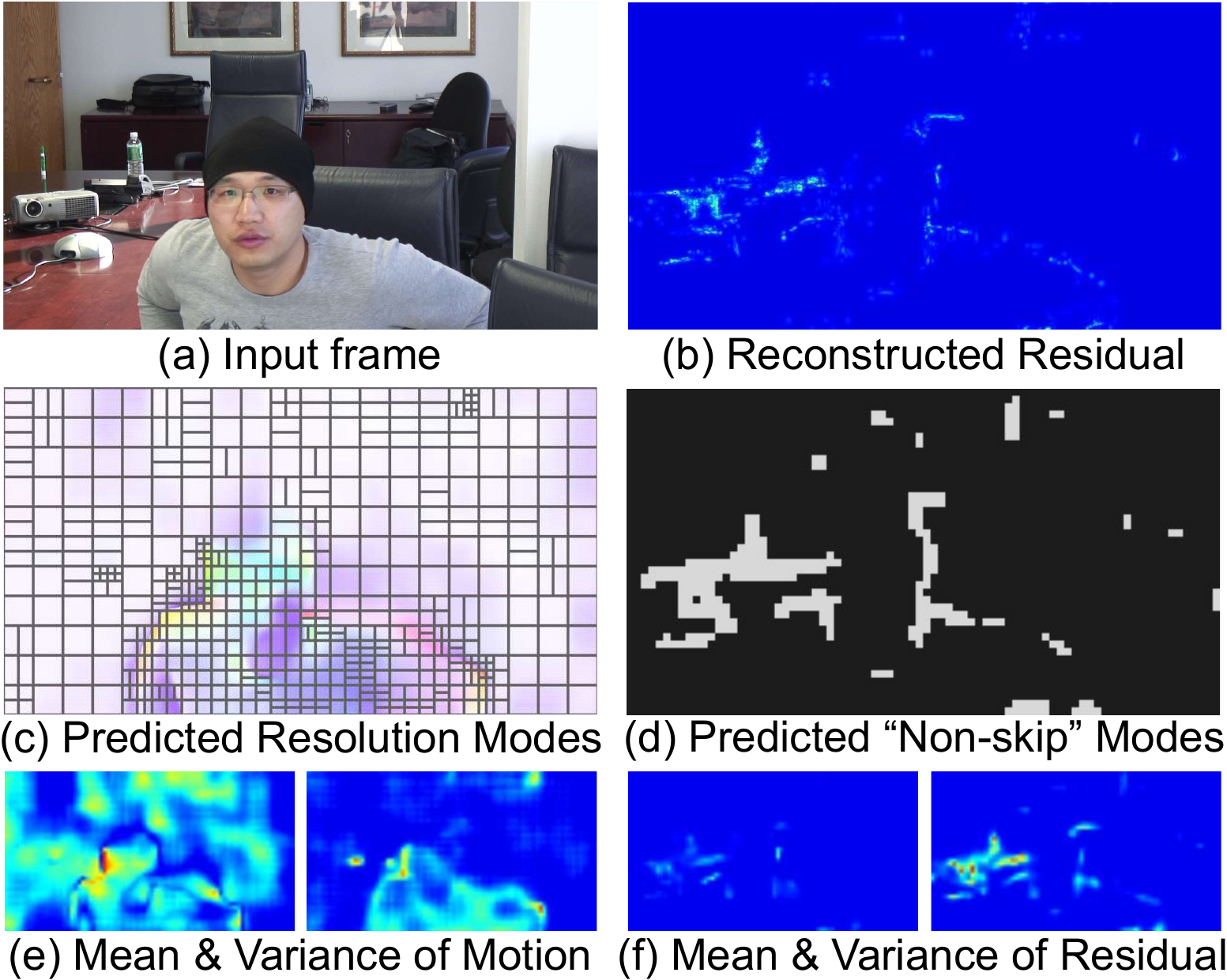}
  \vspace{-2mm}
  \caption{Visualization of the predicted modes for the first P frame (the input frame and the reconstructed residual are shown in (a) and (b)) of the 3rd video from the HEVC Class E dataset by using our proposed methods HAMC (c) and HARC (d). The corresponding mean and variance values (e,f) from the hyperprior networks are also shown. In (b,e,f), the red color (\textit{resp.}, the blue color) denotes large (\textit{resp.}, small) residual/mean/variance value at the corresponding location. In (d), the white color denotes the predicted ``non-skip" mode.\vspace{-6mm}} 
\label{fig:hyper}
\end{figure}

\textbf{Visualization of the Predicted Modes.}
In Fig.~\ref{fig:hyper}, we take the first P frame of the 3rd video from the HEVC Class E dataset as an example and visualize the reconstructed offset and residual maps, as well as their corresponding predicted resolution and ``non-skip" modes. We also visualize their corresponding mean and variance values from the hyperprior networks in HAMC and HARC, respectively. From Fig.~\ref{fig:hyper}(e), we observe that the areas around moving object boundaries contain large mean and/or variance values. Based on such discriminant hyperprior information, our method HAMC will use the small block sizes when encoding motion information in such areas (see Fig.~\ref{fig:hyper}(c)). In contrast, larger block sizes will be preferred for the background areas with small mean and/or variance values. In Fig.~\ref{fig:hyper}(b), a few areas contain significant residual information (see the large mean or variance values in Fig.~\ref{fig:hyper}(f)).
Based on hyperprior information, our HARC method prefers the “non-skip” mode for these blocks when encoding residual information (see Fig.~\ref{fig:hyper}(d)).

\textbf{Running Speed.}
We report the inference speed of our proposed framework based on the videos with the resolution of $1920\times1080$ on the machine with a single 2080TI GPU. 
Our proposed C2F+HAMC+HARC method runs at 3.41fps while our basic C2F framework without adopting both HAMC and HARC methods runs at 3.43fps, which demonstrate that our proposed hyperprior-guided mode prediction methods bring negligible extra computation cost.

\vspace{-2mm}
\section{Conclusion}
\vspace{-1mm}
In this work, we have proposed a new coarse-to-fine (C2F) video compression framework equipped with two newly proposed hyperprior-guided mode prediction schemes HAMC and HARC, which are respectively used for more accurate motion compensation and for compressing motion and residual information with less bit cost. Comprehensive experiments demonstrate our method achieves comparable performance with H.265(HM) in terms of PSNR and generally outperforms VTM in terms of MS-SSIM. Considering that the mode selection strategy is widely used in the conventional codecs like H.265, our work opens a new door for the subsequent researchers to use/extend our hyperprior-guided method to decide other types of optimal modes for better video compression performance.

\noindent\textbf{Acknowledgement}
This work was supported by the National Key Research and Development Project of China (No. 2018AAA0101900) and in part by the National Natural Science Foundation
of China under Grant 62102024.

{\small
\bibliographystyle{ieee_fullname}
\bibliography{egbib}
}

\clearpage


\renewcommand\thesection{\Alph{section}}
\renewcommand{\thefigure}{S\arabic{figure}}

\renewcommand{\thetable}{S\arabic{table}}
\setcounter{section}{0}
\setcounter{figure}{0}
\setcounter{table}{0}




\begin{figure*}
    \includegraphics[width=\textwidth]{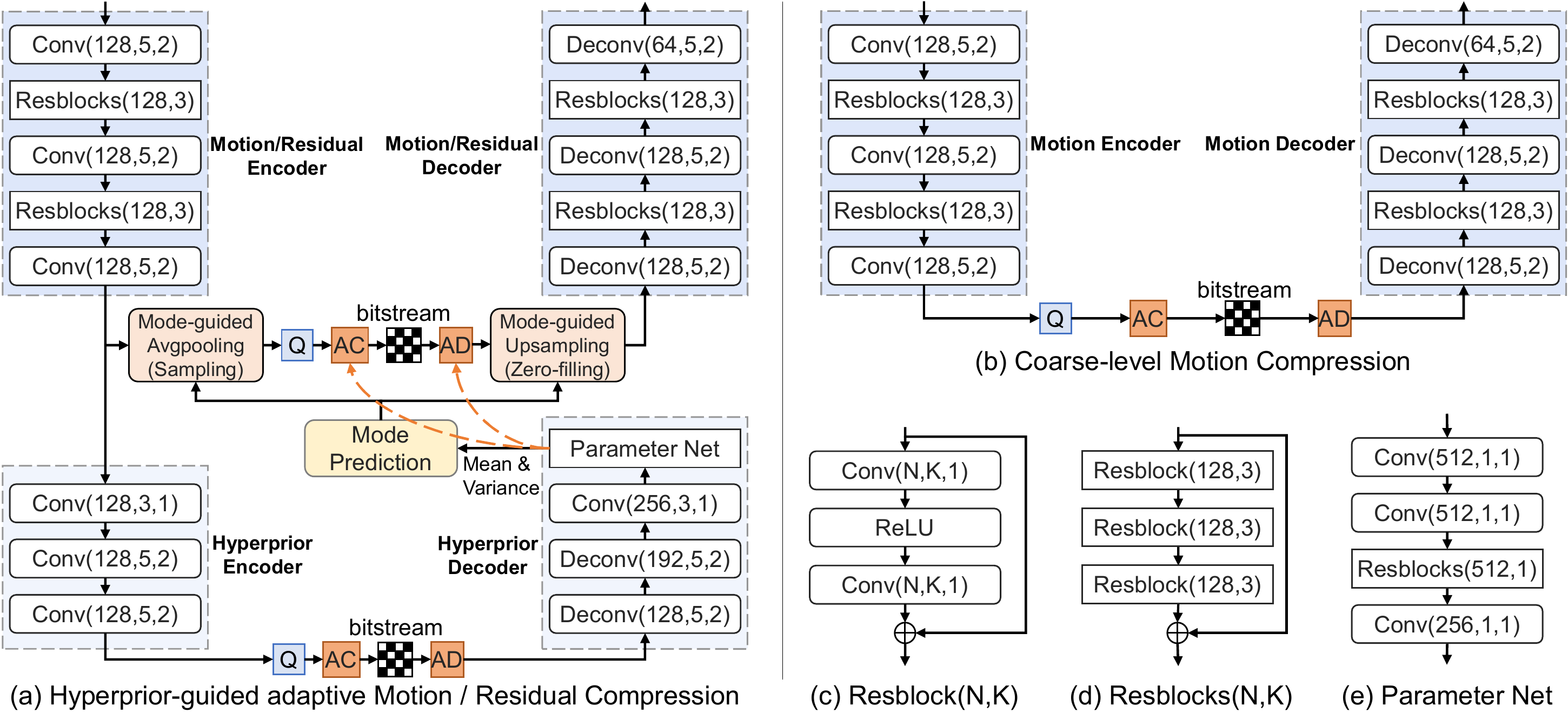}
    \caption{More detailed network structure of (a) the hyperprior-guided adaptive motion and residual compression network, (b) the coarse-level motion compression network and some basic modules including (c) Resblock, (d) Resblocks and (e) the parameter net in the hyperprior decoder. ``Conv(N,K,S)" and ``Deconv(N,K,S)" denote a convolution or a deconvolution layer with the output channel $N$, the kernel size $K\times K$ and the stride $S$.}
    \label{fig:network}
    \vspace{+5mm}
\end{figure*}


\section{More Detailed Network Structure of the Compression Networks}
In Fig.~\ref{fig:network}, we provide more detailed network structure of our hyperprior-guided adaptive motion compression (HAMC) module (see Fig.~\ref{fig:network}(a)), the hyperprior-guided adaptive residual compression (HARC) module (see Fig.~\ref{fig:network}(a)) and the coarse-level motion compression module (see Fig.~\ref{fig:network}(b)). In HAMC, we use the ``mode-guided avgpooling" and ``mode-guided upsampling" for different block resolutions, while in HARC we use the ``mode-guided sampling" and ``mode-guided zero-filling" for the ``skip"/``non-skip" mode. For estimating the bit-rate cost when encoding the quantized features from the hyperprior encoder and the coarse-level motion encoder, we adopt the bit-rate estimation network proposed in \cite{balle2016end}.

\begin{table*}[ht]
    \begin{center}
            \caption{BDBR(\%)~\cite{bjontegaard2001calculation} and BD-PSNR(dB) results of FVC~\cite{hu2021fvc}, ELF-VC~\cite{rippel2021elf}, DCVC~\cite{li2021deep}, our baseline method FVC(re-imp) and our proposed method on different datasets when using the conventional standard H.265(HM)~\cite{sullivan2012overview} as the anchor method. Negative values in BDBR represent bit-rate savings.}\setlength{\tabcolsep}{1.6mm}{
    \begin{tabular}{|l|c|c|c|c|c|c|c|c|c|c|}
    \hline
                            & \multicolumn{5}{c|}{BDBR}                                 & \multicolumn{5}{c|}{BD-PSNR}                               \\
    \hline
                            & FVC    & ELF-VC & DCVC    & FVC(re-imp) & Ours            & FVC    & ELF-VC  & DCVC    & FVC(re-imp) & Ours    \\
    \hline
        HEVC Class A        & -      &   -    & -       & -8.83       & \textbf{-18.64} & -      &   -     & -       & 0.27        & \textbf{0.57}   \\
    \hline
        HEVC Class B        & 42.70  &   -    & 13.63   & -4.57       & \textbf{-19.91} & -0.58  &   -     & -0.28   & 0.09        & \textbf{0.37}   \\
    \hline
        HEVC Class C        & 44.80  &   -    & 49.21   & 26.61       & \textbf{12.32}  & -1.29  &   -     & -1.37   & -0.83       & \textbf{-0.42}  \\
    \hline
        HEVC Class D        & 38.45  &   -    & 25.67   & 14.63       & \textbf{-1.21}  & -1.25  &   -     & -0.92   & -0.54       & \textbf{0.04}   \\
    \hline
        HEVC Class E        & 119.76 &   -    & 117.00  & 37.63       & \textbf{-9.53}  & -1.69  &   -     & -1.86   & -0.55       & \textbf{0.25}   \\
    \hline
        Average of HEVC     & 61.43  &   -    & 51.38   & 18.58       & \textbf{-7.39}  & -1.20  &   -     & -1.11   & -0.45       & \textbf{0.16}   \\
    \hline
        UVG                 & 58.10  & 23.89  & 46.51   & 19.59       & \textbf{-9.23}  & -1.06  & -0.49   & -0.89   & -0.38       & \textbf{0.21}   \\
    \hline
        MCL-JCV             & 36.22  & 19.48  & 29.48   & 20.78       & \textbf{-3.51}  & -0.70  & -0.45   & -0.64   & -0.43       & \textbf{0.09}   \\
    \hline
    \end{tabular}}
    \label{tab:BDBRPSNR}
    \end{center}
\end{table*}

\begin{figure}[t]
  \centering
  \begin{minipage}[c]{0.5\linewidth}
  \centering
    \includegraphics[width=\textwidth]{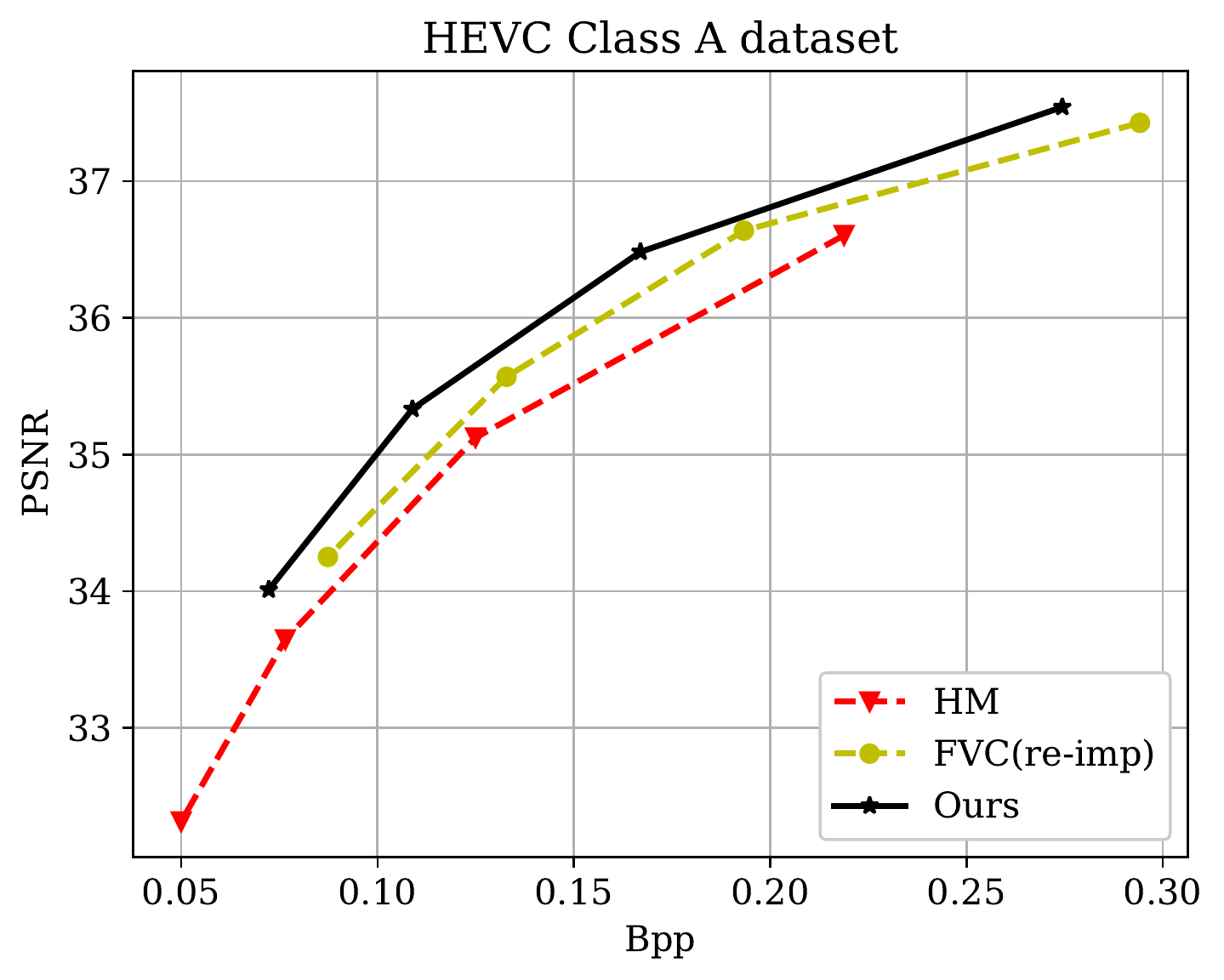}
  \end{minipage}%
  \begin{minipage}[c]{0.5\linewidth}
  \centering
    \includegraphics[width=\textwidth]{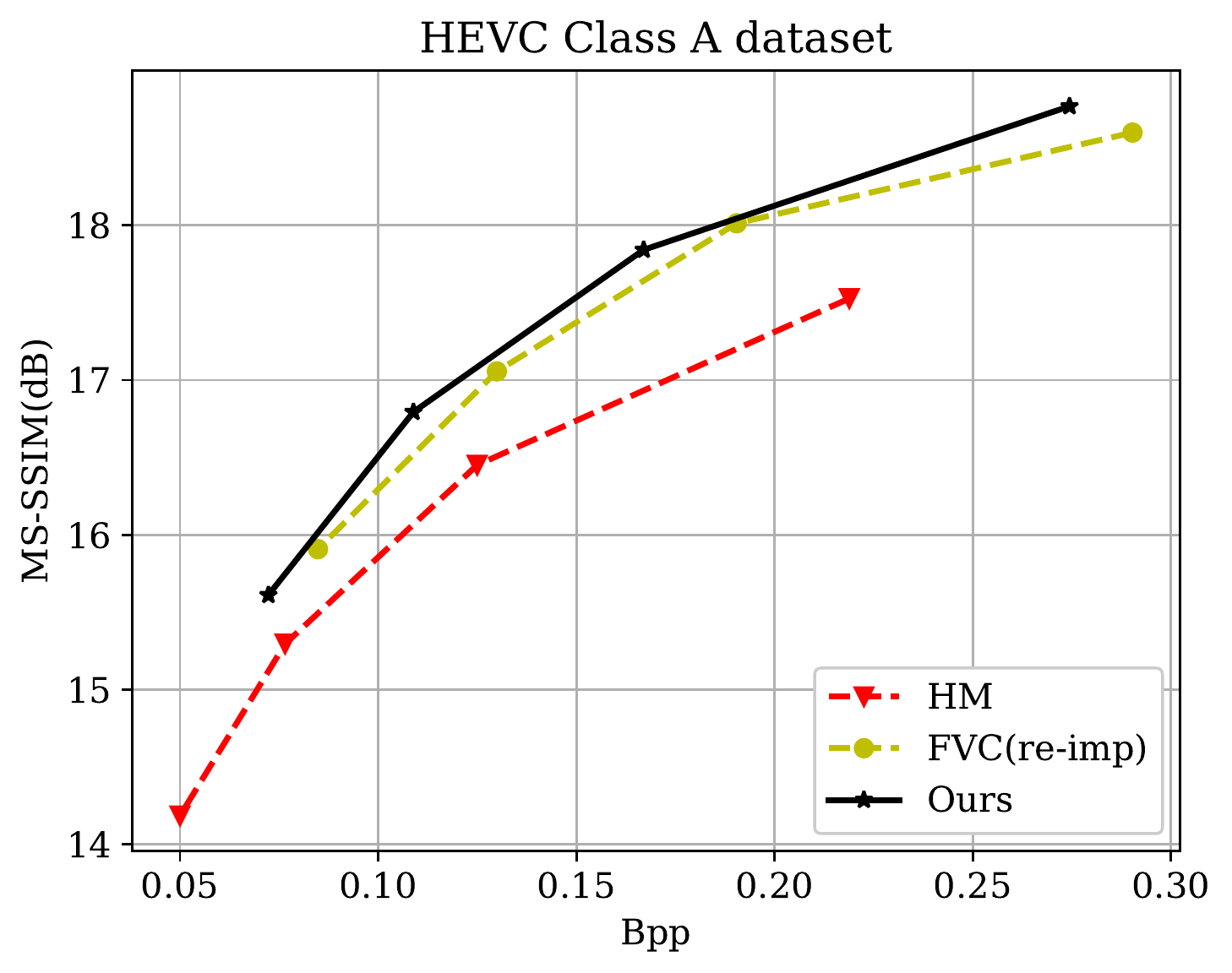}
  \end{minipage}
  \vspace{-2mm}
    \caption{The experimental results on the HEVC Class A dataset.\vspace{-2mm}}
  \label{fig:aresult}
\end{figure}

\section{Experimental Results on the HEVC Class A dataset}
As mentioned in our main paper, we further report the experimental results on the HEVC Class A dataset in Fig.~\ref{fig:aresult}. Specifically, we evaluate our proposed method on two video sequences \textit{Traffic} and \textit{PeopleOnStreet} with the resolution of $2560\times1600$ from the HEVC Class A dataset. We observe that our method achieves 0.7dB improvement when compared with the standard H.265(HM), which demonstrates that our method can also achieve excellent compression performance on high resolution video sequences.

\section{BDBR Results}
In Table~\ref{tab:BDBRPSNR}, we also report the detailed BDBR\cite{bjontegaard2001calculation} and BD-PSNR results on multiple datasets when compared with the conventional standard H.265(HM). It is observed that our method outperforms all other learning based methods by a large margin. When compared with H.265(HM), our method achieves better results on all high-resolution video datasets (\ie, HEVC Class A,B,E, UVG and MCL-JCV datasets).

\begin{figure*}[t]
  \begin{minipage}[c]{0.33\textwidth}
  \centering
    \includegraphics[width=\textwidth]{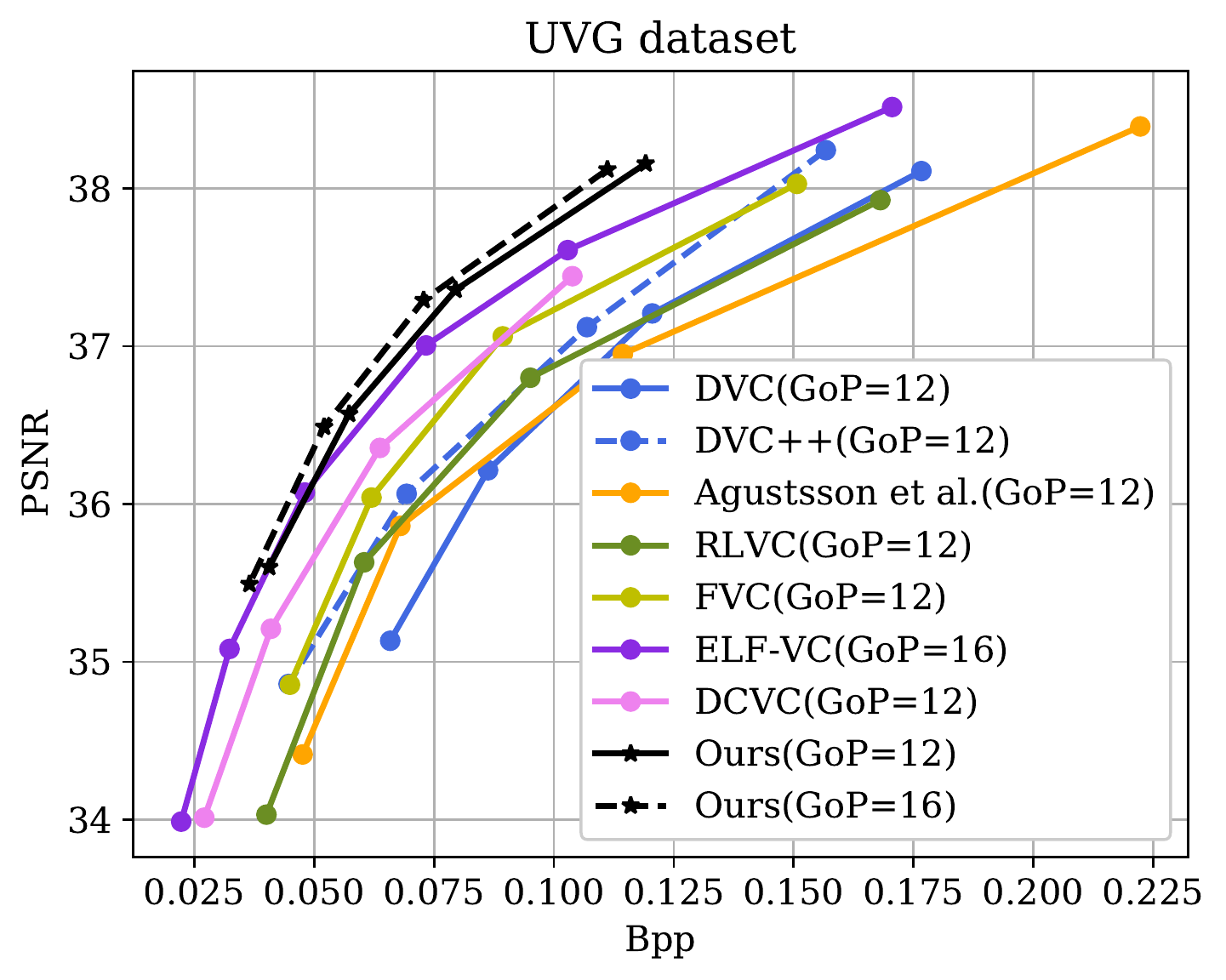}
  \end{minipage}%
  \begin{minipage}[c]{0.33\textwidth}
  \centering
    \includegraphics[width=\textwidth]{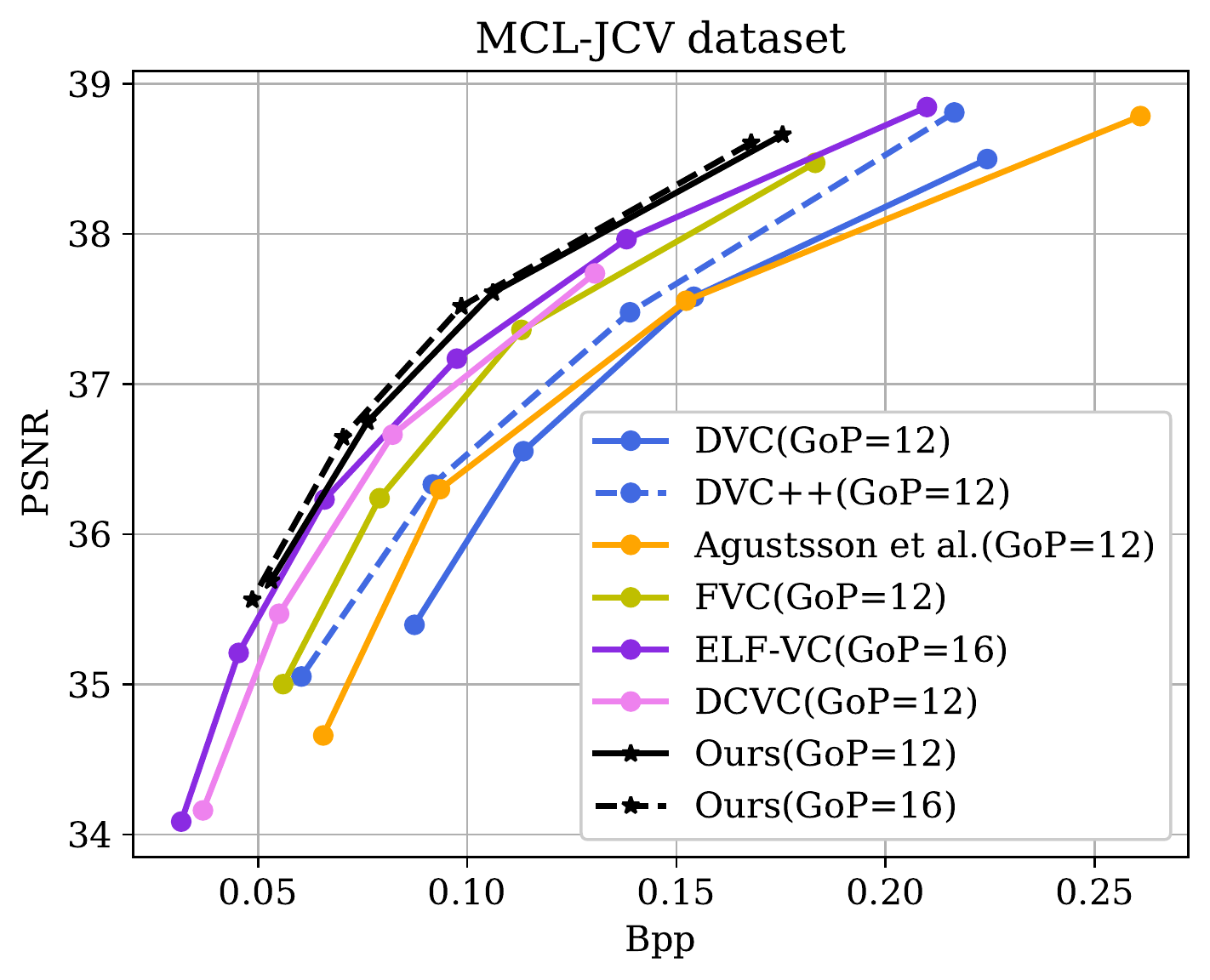}
  \end{minipage}%
  \begin{minipage}[c]{0.33\textwidth}
  \centering
    \includegraphics[width=\textwidth]{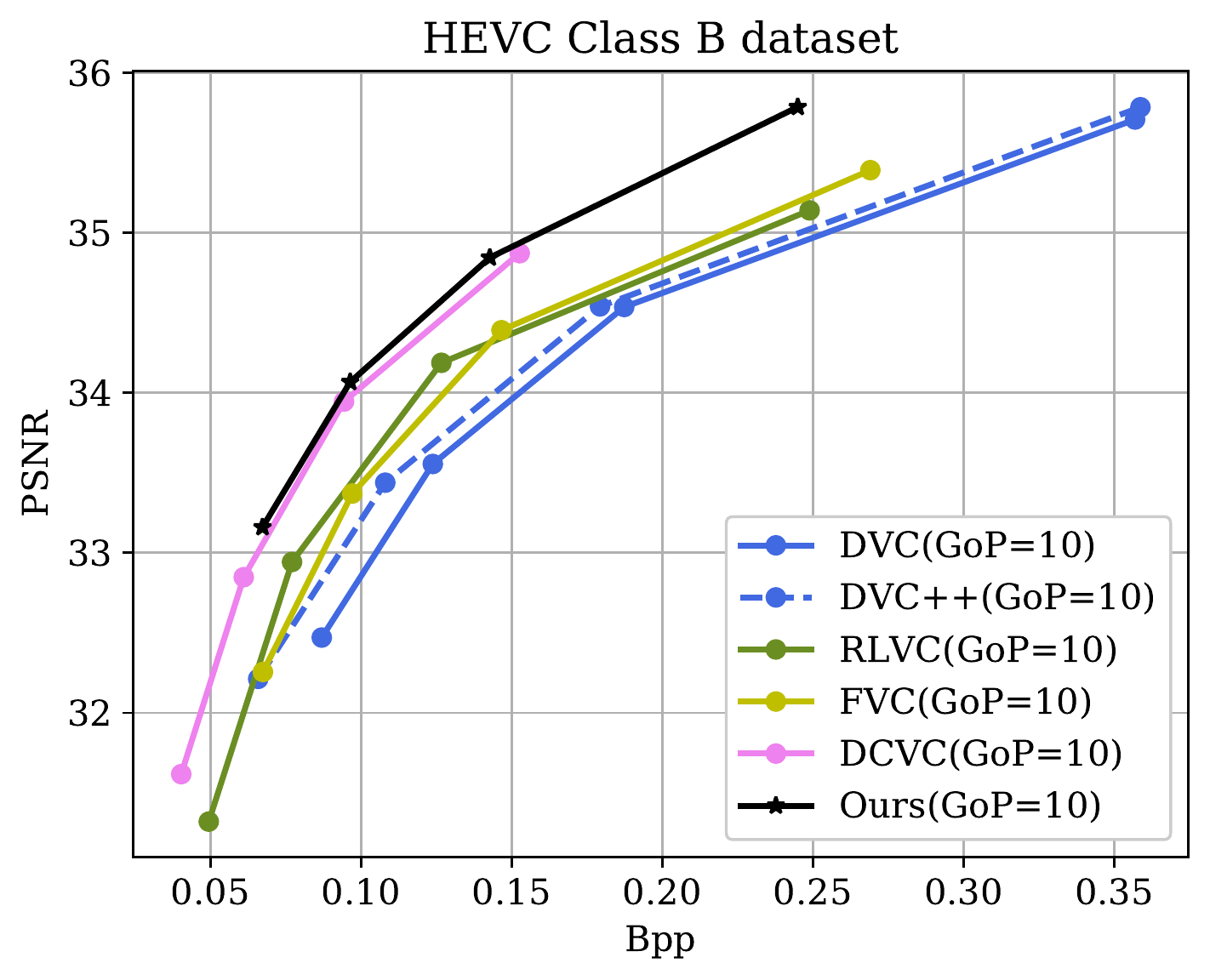}
  \end{minipage}
  \begin{minipage}[c]{0.33\textwidth}
  \centering
    \includegraphics[width=\textwidth]{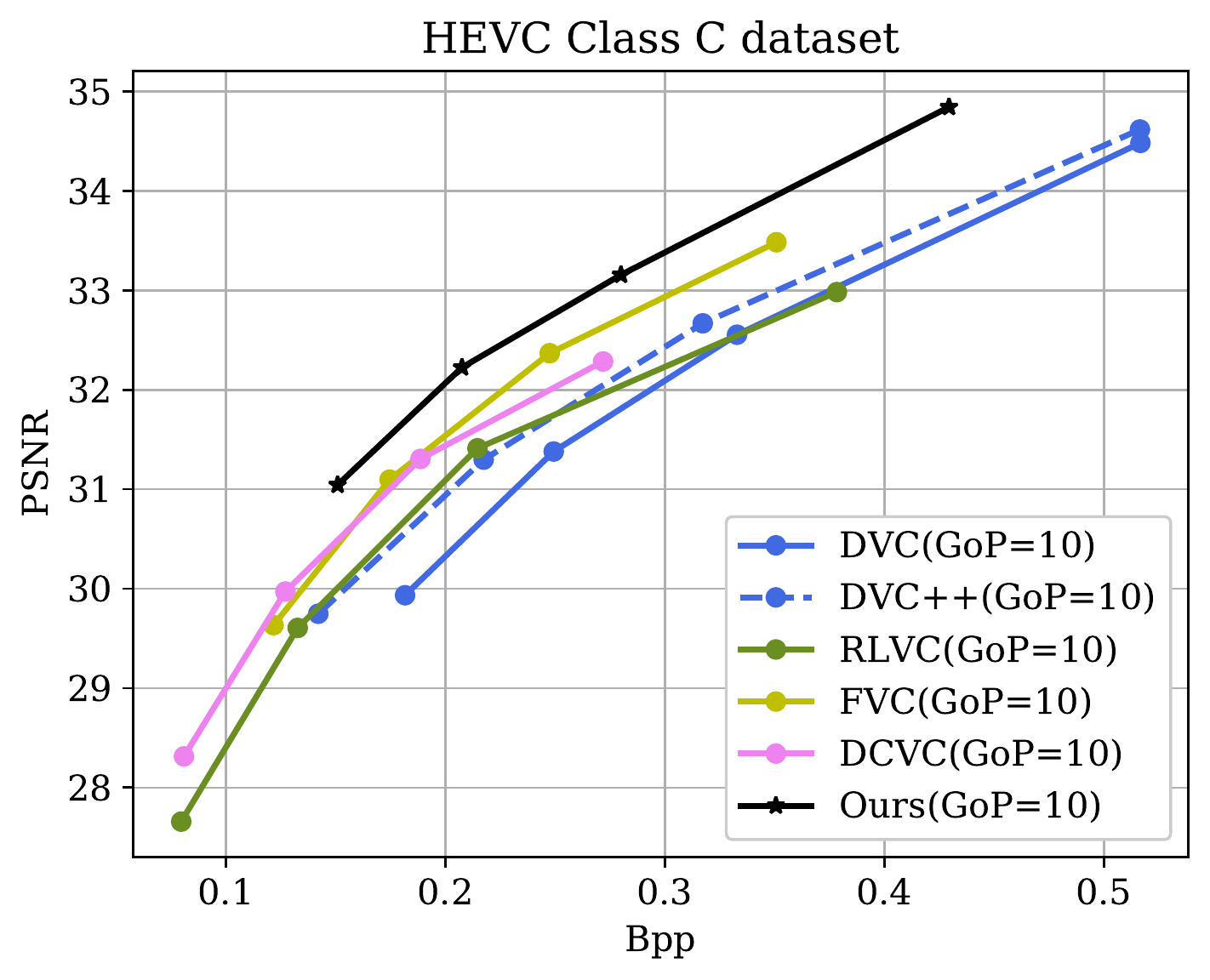}
  \end{minipage}%
  \begin{minipage}[c]{0.33\textwidth}
  \centering
    \includegraphics[width=\textwidth]{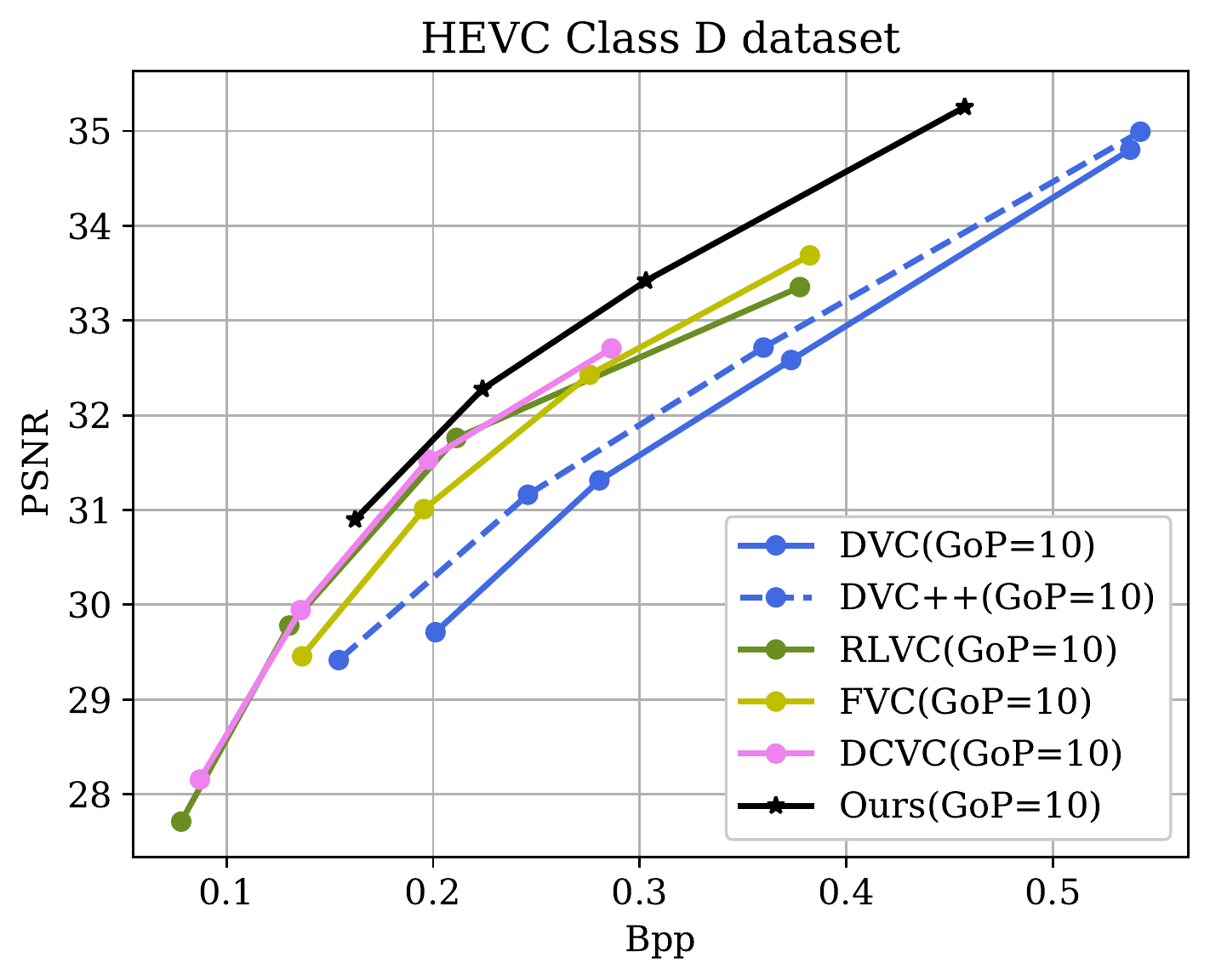}
  \end{minipage}%
  \begin{minipage}[c]{0.33\textwidth}
  \centering
    \includegraphics[width=\textwidth]{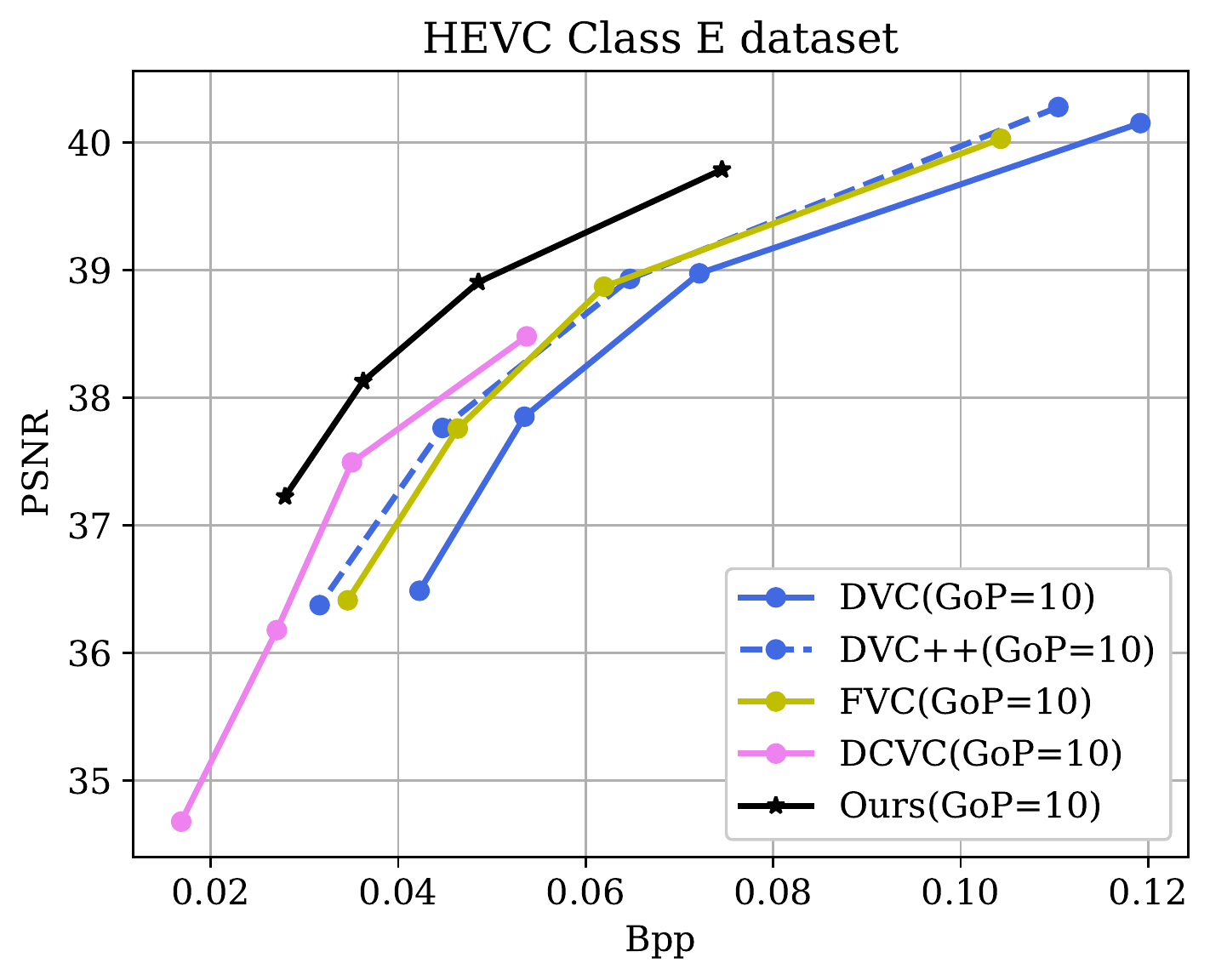}
  \end{minipage}%
    \caption{The PSNR results of different methods on the UVG, MCL-JCV, HEVC Class B, Class C, Class D and Class E datasets under the new settings with small GoP sizes (\ie, GoP=10, 12 or 16).}
  \label{fig:gop10result}
\end{figure*}

\section{GoP Size}
In our main paper, we set the GoP size as 100 to fairly compare our work with the conventional standard H.265(HM)~\cite{sullivan2012overview} and VTM~\cite{sullivan2020versatile}. In order to minimize the mismatch between the short training sequence and long testing sequence, we also follow the conventional methods to use a model with larger $\lambda$ value at the fourth frame of each four consecutive P frames to reduce the cumulative error. 

To fairly compare our method with the previous learning-based methods with small GoP sizes, In Fig.~\ref{fig:gop10result}, we also evaluate our method under their settings. We observe that our method achieves better performance when compared with all the previous learning-based deep video compression methods. For example, when compared with the recently proposed ELF-VC~\cite{rippel2021elf}, our method based on the same GoP size (\ie, GoP=16) achieves 0.3dB improvement at 0.1bpp on the UVG dataset.

In Fig.~\ref{fig:gop10result}, We further provide the results of ``DVC++", which incorporates our newly proposed methods HAMC and HARC into the existing DVC framework. It is observed that DVC++ achieves 0.5dB improvement at 0.1bpp on the MCL-JCV dataset when compared with the baseline method DVC, which further demonstrates our proposed hyperprior-guided mode prediction methods are general and can be readily used/combined with other deep video compression methods like DVC.

\begin{figure*}[t]
  \centering
  \centering
    \includegraphics[width=\linewidth]{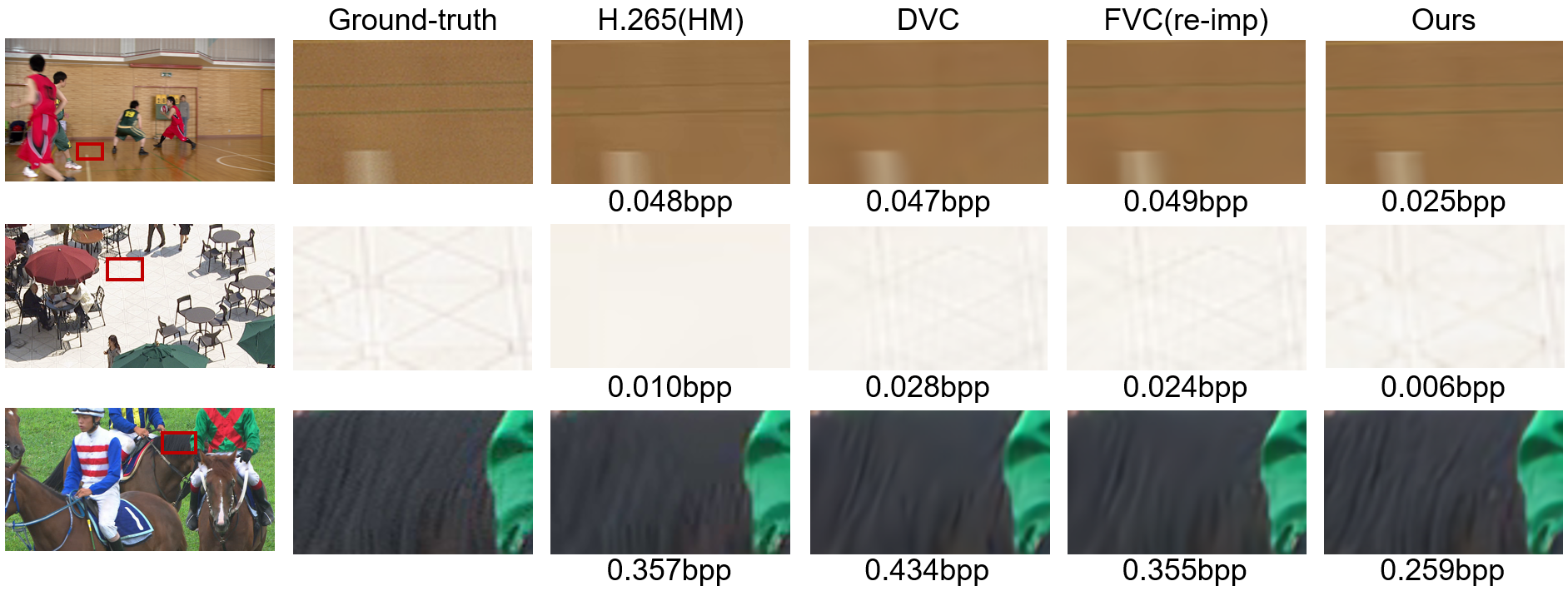}
    \caption{Visualization results from H.265(HM)~\cite{sullivan2012overview}, DVC~\cite{lu2019dvc}, the re-implemented baseline method FVC(re-imp)~\cite{hu2021fvc} and our proposed method.}
  \label{fig:visualization}
\end{figure*}

\section{Visualization}
We provide the visualization results from different methods in Fig.~\ref{fig:visualization}. The existing methods including the conventional method H.265(HM) and other learning based methods DVC and FVC(re-imp) sometimes produce wrong colors or lose the details, which leads to worse reconstruction quality. For example, in the first row of Fig.~\ref{fig:visualization}, the two lines produced by H.265(HM) has the wrong color and the first reconstructed line (\ie, the upper one) by using DVC and FVC(re-imp) is also blurred, while our method can better reconstruct the two lines with the right color. Additionally, H.265(HM), DVC and FVC(re-imp) lose much details and thus produce blurred reconstruction results, while our method with the least bit-rate cost achieves the best reconstruction quality and our results contain much more detailed textures (see the left region of the second row and the horse hair from the third row in Fig.~\ref{fig:visualization}).  

\section{Setting of H.265(HM) and VTM}
When evaluating the performance of the conventional methods H.265(HM) and VTM, we directly use the official code with the \textit{low delay P} setting. The version of H.265(HM) and VTM are 16.20 and 11.2, respectively. Additionally, for both H.265(HM) and VTM, we use $22, 25, 28, 31$ as the QP values for generating the results.


\end{document}